\documentclass[final,5p,times,twocolumn,nopreprintline]{elsarticle}

\usepackage{amssymb}

\usepackage{lineno}

\usepackage[table, dvipsnames]{xcolor}
\usepackage{soul}
\usepackage{amsmath}
\usepackage{gensymb}
\usepackage{bm}
\usepackage[ruled,linesnumbered,noend,noline]{algorithm2e}
\usepackage{tabularx}
\usepackage{caption}
\usepackage{subcaption}
\usepackage{multirow}
\usepackage{dblfloatfix}
\usepackage{hyperref}
\usepackage{pgfplots}
\pgfplotsset{compat=1.17}

\journal{Elsevier}

\begin{document}
\switchlinenumbers
\begin{frontmatter}



\title{Urban GeoBIM construction by integrating semantic LiDAR point clouds with as-designed BIM models}


\author[polyu,polyusri]{Jie Shao}
\ead{jie.shao@polyu.edu.hk}

\author[polyu,polyusri,polyuscri]{Wei Yao\corref{cor1}}
\ead{wei.hn.yao@polyu.edu.hk}

\author[polyu]{Puzuo Wang\corref{cor1}}
\ead{puzuo.wang@connect.polyu.hk}

\author[polyu]{Zhiyi He}

\author[CAS]{Lei Luo}

\address[polyu]{Dept. of Land Surveying and Geo-Informatics, The Hong Kong Polytechnic University, Hung Hom, Kowloon, Hong Kong}
\address[polyusri]{The Hong Kong Polytechnic University Shenzhen Research Institute, Shenzhen, China}
\address[polyuscri]{Otto Poon Charitable Foundation Smart Cities Research Institute, The Hong Kong Polytechnic University, Hung Hom, Hong Kong}
\address[CAS]{Key Laboratory of Digital Earth Science, Aerospace Information Research Institute, Chinese Academy of Sciences, Beijing, China}

\cortext[cor1]{Corresponding authors.}


\begin{abstract}
Developments in three-dimensional real worlds promote the integration of geoinformation and building information models (BIM) known as GeoBIM in urban construction. Light detection and ranging (LiDAR) integrated with global navigation satellite systems can provide geo-referenced spatial information. However, constructing detailed urban GeoBIM poses challenges in terms of LiDAR data quality. BIM models designed from software are rich in geometrical information but often lack accurate geo-referenced locations. In this paper, we propose a complementary strategy that integrates LiDAR point clouds with as-designed BIM models for reconstructing urban scenes. A state-of-the-art deep learning framework and graph theory are first combined for LiDAR point cloud segmentation. A coarse-to-fine matching program is then developed to integrate object point clouds with corresponding BIM models. Results show the overall segmentation accuracy of LiDAR datasets reaches up to 90\%, and average positioning accuracies of BIM models are 0.023 m for pole-like objects and 0.156 m for buildings, demonstrating the effectiveness of the method in segmentation and matching processes. This work offers a practical solution for rapid and accurate urban GeoBIM construction.

\end{abstract}




\end{frontmatter}


\section{Introduction}
\label{sec:intro}

Urban scene is a dynamic environment that comprises buildings, pole-like objects, and other infrastructures, making large-scale three-dimensional (3D) urban construction crucial for various applications such as urban design, transportation, and tourism \cite{9524478}. With advancements in information technology, an increasing number of technologies for reconstructing urban scenes are being proposed, with digital twins and the 3D real scenes being the current research hotspots in the field \citep{WANG2022100351,YANG2022107804}. These technologies describe physical entities in the real world into virtual spaces through digitization \citep{9382113}. By leveraging digitized entities, urban infrastructures can be accurately understood and provide valuable references for urban management and decision-making.

Building information modeling (BIM) is a widely used digital tool for urban scene construction, integrating geometry, attributes, and status information throughout the whole life cycle of infrastructure, from engineering design to construction and operation \citep{LOTFI2021102559}. Nowadays, there are numerous software and tools, such as Autodesk Revit, related to BIM technologies that have been developed and applied in 3D urban construction for various purposes, such as hazard analysis \citep{YANG2021103626} and environmental impact assessment \citep{FERNANDEZRODRIGUEZ2018494}. However, BIM models created in software manually are often represented based on a local coordinate system, lacking accurate geo-referenced locations, which poses challenges for accurate urban construction \citep{UGGLA2018554}. In recent years, some engineering projects and academic studies have highlighted the advantage of using geo-referenced BIM models in various fields, such as urban planning and asset management \citep{ijgi6020053,DIAKITE2020101453}. As a result, the integration of geoinformation and BIM models, known as GeoBIM, is gaining increasing attention within the urban construction community \citep{doi:10.1080/14498596.2019.1627253}.

As mentioned earlier, BIM models for infrastructures are typically endowed with geoinformation through manual input using human-computer interaction. However, this approach can be time-consuming and less accurate compared to using 3D real scene data. To address this issue, some studies have proposed inverse procedural modeling approaches \citep{10.1145/2732527}, where BIM models are generated from geographic information system (GIS) data. Light detection and ranging (LiDAR) integrated with a global navigation satellite system (GNSS) is a powerful tool that can directly provide accurate geospatial GIS data, i.e., geo-referenced 3D point cloud \citep{FERNANDEZALVARADO2022104251}. By leveraging modeling technology, LiDAR point clouds can be meshed to reconstruct multiple levels of detail (LoD) urban models \citep{rs14030618,BOLOURIAN2020103250}. However, existing methods for LiDAR-based urban construction often have strict requirements on data quality, including accuracy and integrity. Single-platform LiDAR systems may struggle to provide complete spatial information for complex urban scenes, and challenges may arise in acquiring data from difficult-to-reach observation locations or fusing data from multi-platform LiDAR systems, resulting in inaccurate and incomplete urban models. Additionally, most previous studies have focused on simple-shaped entities, with limited attention to complex entities \citep{WANG2022103997}. Therefore, there is a need for further exploration of efficient and high-quality urban construction.

This study aims to address the research gap by proposing a rapid and accurate method for urban GeoBIM construction. The method integrates semantic LiDAR point cloud with as-designed BIM models to leverage the strengths of both data sources. The LiDAR point cloud provides accurate geo-referenced information, while BIM models provide geometric and attribute information. The proposed method will reconstruct urban scenes even when high-quality LiDAR data or complex modeling technologies are not available. Specifically, LiDAR point clouds and as-designed infrastructure BIM models are first collected for study areas. A state-of-the-art deep learning framework is then used for semantic segmentation of LiDAR point clouds, in which an adaptive extraction method is used for automated labeling of training samples and to increase efficiency; moreover, a graph-based clustering approach combining multiple geometric attributes is adopted to instance and optimize segmentation of infrastructures, which has the potential to improve the segmentation accuracy. Next, a coarse-to-fine matching strategy is developed to integrate LiDAR point clouds with the corresponding as-designed BIM models, especially proposing an automated approach for coarse alignment of BIM models with sparse point clouds and LiDAR data of buildings through non-planar 4-point pairs. Following the introduction section, Section~\ref{sec:review} presents a comprehensive literature review of the related works on point cloud segmentation and matching of GIS data and BIM models. The proposed method will be described in detail in Section~\ref{sec:method}. Experimental results will be introduced in Section~\ref{sec:res} and discussed in Section~\ref{sec:discuss}, and then conclusions are presented.

\section{Literature review}
\label{sec:review}

\subsection{3D point cloud segmentation}
\label{sec:review_pcss}

The basic premise of this study is to segment individual objects from urban scenes and provide geo-referenced data sources for the corresponding BIM models of each infrastructure. Conventional approaches, such as region growing \citep{DIMITROV201532} and DBSCAN \citep{WANG2016170}, typically segment point clouds based on geometric characteristics, such as planar and cylinder shapes. As a result, these methods are generally suitable for simple objects with regular geometric shapes \citep{WANG2022103997} that can be represented by fixed parameters. However, complex-shaped objects are difficult to segment based on single geometric features, as they often require cumbersome rule-based designs, are sensitive to data sources, and lack generalization capabilities. While there are some methods that can represent complex geometric shape features \citep{5152473,4563023}, obtaining accurate individual objects solely through traditional segmentation methods in urban scenes remains difficult.

Recent advancements in machine learning and deep learning technologies have enabled more intelligent extraction of rich semantic information from point clouds. Traditional machine learning methods often require a manual definition of geometry or topology features, followed by the use of classifiers such as support vector machine and random forest to label points \citep{GUO201571,WANG201975}. In contrast, deep learning technologies can automatically construct classification models and classify objects by directly inputting point clouds into a deep neural network, without the need for handcrafted geometry or topology features. This has become the dominant method in current point cloud classification \citep{9127813}, with the introduction of the PointNet \citep{8099499} framework setting a precedent in this field. The PointNet framework addresses challenges such as point cloud disorder and permutation and rotation invariances with low cost. However, it ignores the relationship between neighboring points in the local field. To overcome this limitation, an expanded framework called PointNet++ \citep{10.5555/3295222.3295263} was proposed, which incorporates local hierarchical modules to improve segmentation accuracy by considering features in local and global receptive fields. Inspired by these deep learning frameworks, several other networks with different strategies have been proposed for 3D point cloud segmentation, including convolutional-based networks such as RS-CNN \citep{8953930} and KPConv \citep{9010002}, graph-based networks such as PointGCN \citep{8462291}, and deep feature embedding \citep{HUANG202062}.

The aforementioned networks commonly perform well in semantic segmentation of point clouds but still struggle with segmenting individual instances, which has led to the recent study of higher-level instance segmentation technologies \citep{WANG2022103997}, such as SGPN \citep{8578370}, PointGroup \citep{9156675}, and OccuSeg \citep{9157103}. These methods provide better references for intelligent segmentation of individual object point clouds. However, these deep learning frameworks are often reliant on manually labeled training samples, which can be a time-consuming process in large-scale scenes. Moreover, most current instance segmentation methods rely on point cloud clustering based on semantic segmentation results, such as graph-based instance clustering strategies \citep{9156675,9157103}. These works typically consider the distance between neighboring points or voxels to segment instances from the same semantic class, resulting in clustering results that are sensitive to the semantic segmentation and distance threshold settings. Incorrect semantic class assignments or improper threshold settings can adversely affect clustering qualities and the final integration of LiDAR data with BIM models.

\begin{figure*}[b!]
\centering
\includegraphics[width=1.0\linewidth]{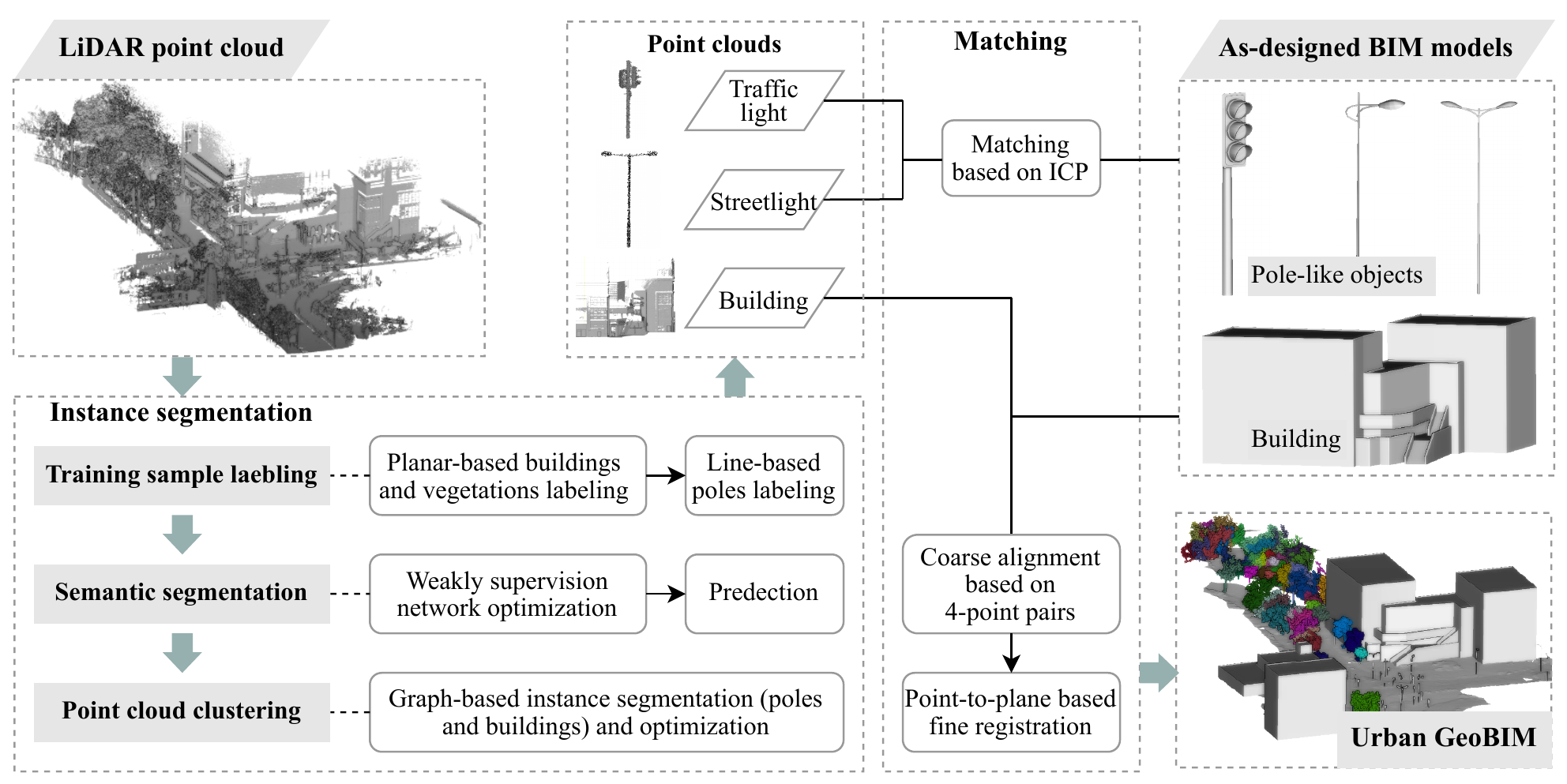}
\caption{Workflow of this paper, including preparation of LiDAR point clouds and as-designed BIM models, instance segmentation of LiDAR point clouds, and coarse alignment and fine registration of instance point clouds and BIM models.}
\label{fig:overview}
\end{figure*}

\subsection{Integration of GIS data and BIM model}
\label{sec:review_gis&bim}

The integration of GIS and BIM aims to bring as-designed BIM models, which are typically in local coordinate systems, into a GIS environment \citep{ijgi6020053}. Conventional approaches involve manual definition of geo-referenced information for BIM models during the design phase using software \citep{ijgi7080311}. However, the accuracy of GIS and BIM integration can be subjective and prone to errors introduced during manual operations. Additionally, existing methods struggle to generate standardized solutions that are scalable for integrating a large number of BIM models into the geo-referenced environment.

In practice, the integration of GIS data, such as LiDAR point clouds, and as-designed BIM models can be addressed through data matching techniques commonly used in computer graphics and remote sensing communities \citep{https://doi.org/10.1111/j.1467-8659.2011.01884.x,TOTH201622}. The core of matching is to establish uniform coordinate systems for different data by accurate correspondences \citep{SHAO2020214,SHAO2022103067}. Typically, the corresponding relationship between two spatial data can be represented by rigid-body transformations involving rotation and translation, where geometric features are often detected to calculate transformation parameters. For instance, some researchers have projected 3D building data into a 2D plane along the z-axis to generate planar features of building footprints for matching 3D GIS data and the corresponding BIM models \citep{DIAKITE2020101453}. These methods perform well for buildings with irregular or unique footprints but face challenges for buildings with similar footprints, in which case 3D features may provide more information for matching. However, there are limited automated approaches and common methods still rely on manual operations \citep{ijgi7080311}. BIM models are typically described by polygon meshes, where each mesh consists of multiple connection points. Hence, the matching of LiDAR data and BIM models can also be considered as point cloud registration.

Currently, there are automated registration algorithms based on feature descriptors that have been proposed for point clouds, such as point feature histograms (PFH) \citep{4650967} and scale-invariant feature transform (SIFT) \citep{10.1023/B:VISI.0000029664.99615.94} algorithms, which are typically suitable for continuously distributed point clouds. However, BIM models are composed of discrete isolated points and may not provide effective feature descriptors for registering with LiDAR point clouds. In contrast, the state-of-the-art algorithm, 4-Points Congruent Sets (4PCS), matches almost-planar 4-point basis in two inputs \citep{10.1145/1399504.1360684,SHAO201916}. This algorithm detects features from a global scale without considering the distribution of local points and can provide a reference for the registration of LiDAR data and BIM models. These algorithms are typically applied to coarse alignment and may not obtain accurate registration results. The conventional algorithm used for fine registration is the Iterative Closest Point (ICP) algorithm, which registers point clouds by minimizing the distance between pairs or corresponding points in two datasets \citep{121791}. However, this method may not be suitable for datasets with low overlap, as it relies on rich structural information. Therefore, further research is needed to develop fine registration for incomplete LiDAR data and BIM models.

\begin{figure*}[b!]
\centering
\begin{subfigure}[b]{0.49\textwidth}
    \centering
    \includegraphics[width=1\linewidth]{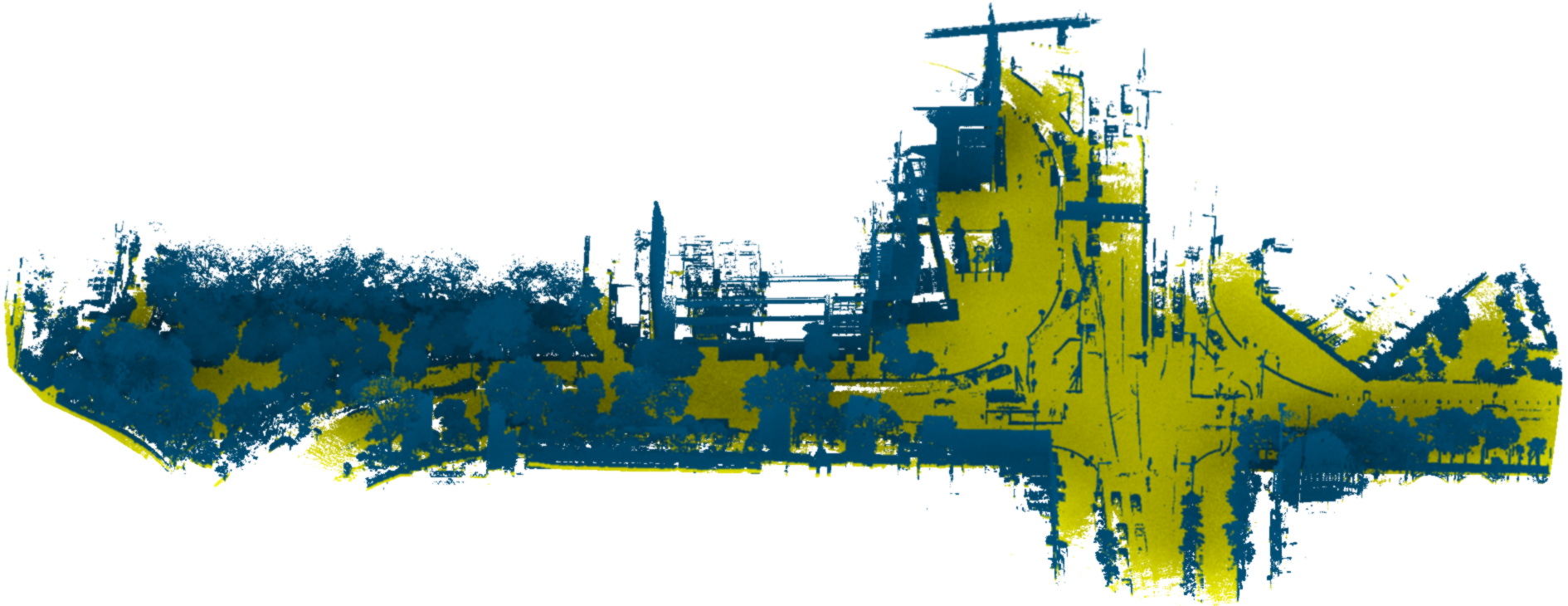}
\caption{}
\end{subfigure}
\hfill
\begin{subfigure}[b]{0.49\textwidth}
    \centering
    \includegraphics[width=1\linewidth]{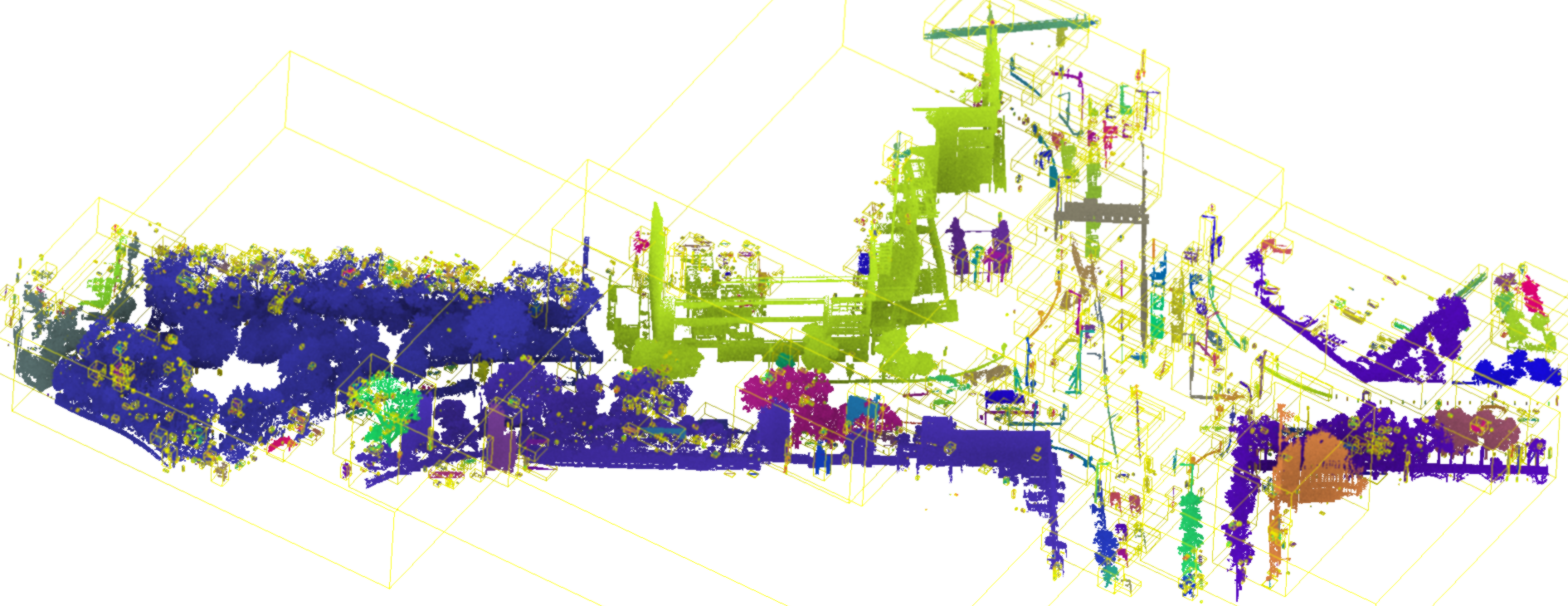}
\caption{}
\end{subfigure}
\caption{Ground filtering (a) and non-ground points clustering (b) of Dataset 1. In (a), yellow points represent the ground, blue points represent non-ground, and random colors in (b) represent different clusters.}
\label{fig:ground_filter}
\end{figure*}

\begin{figure*}[b!]
\centering
\begin{subfigure}[b]{0.3\textwidth}
    \centering
    \includegraphics[width=1\linewidth]{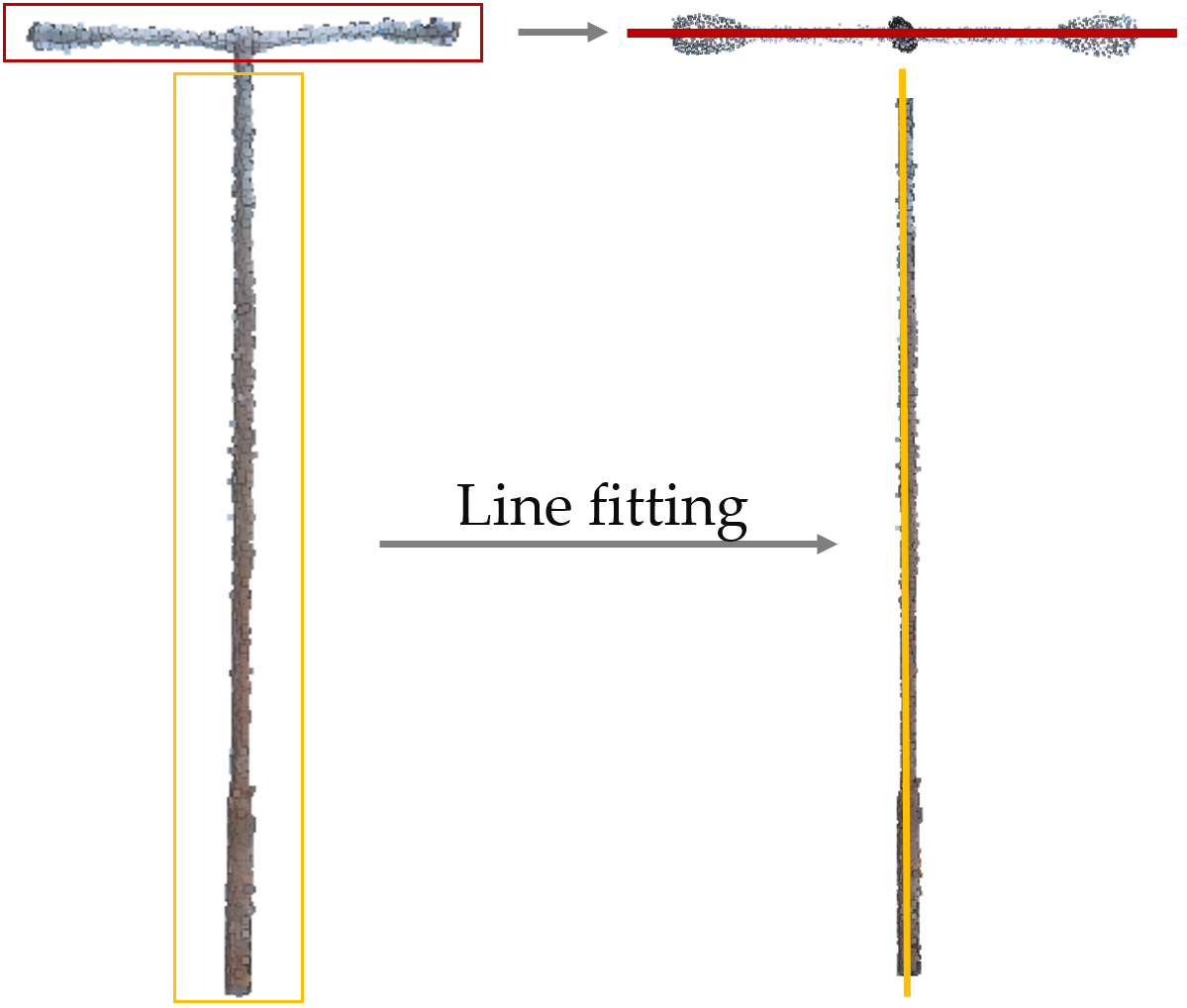}
    \caption{}
    \label{fig:pole_sample}
\end{subfigure}
\hfill
\begin{subfigure}[b]{0.34\textwidth}
    \centering
    \includegraphics[width=1\linewidth]{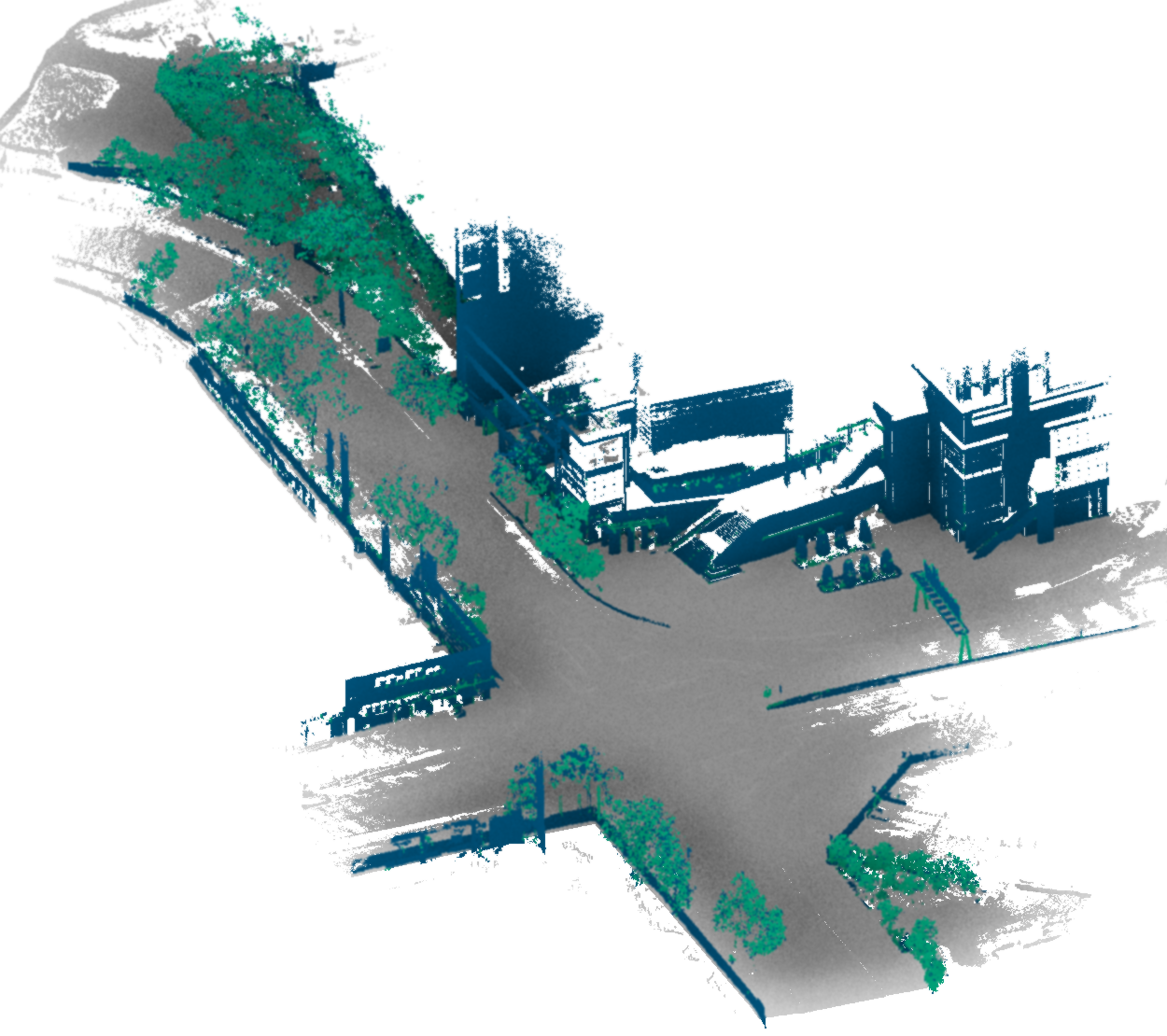}
    \caption{}
    \label{fig:build&veg_label}
\end{subfigure}
\hfill
\begin{subfigure}[b]{0.34\textwidth}
    \centering
    \includegraphics[width=1\linewidth]{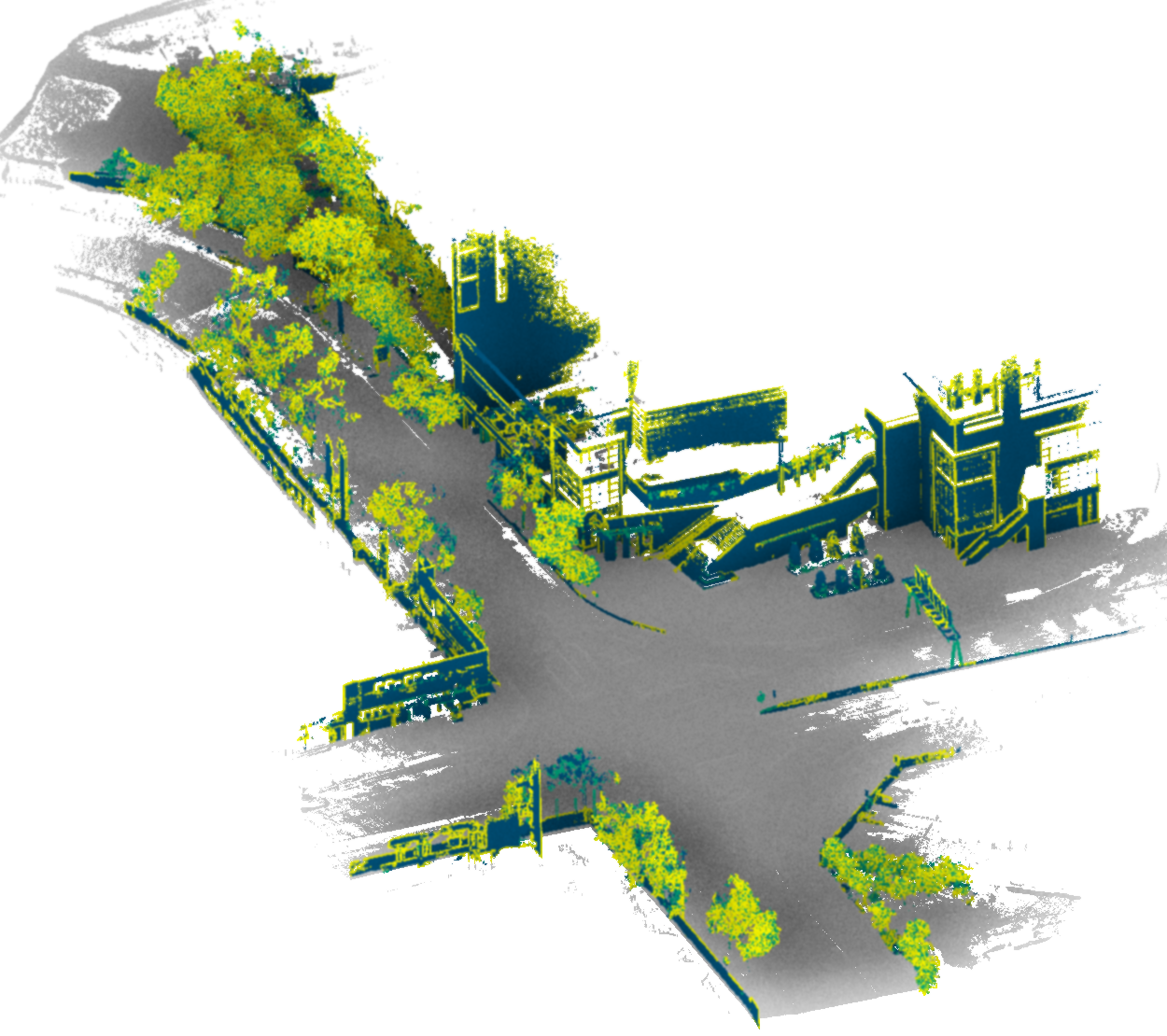}
    \caption{}
    \label{fig:build&veg_voxel_label}
\end{subfigure}
\caption{Training sample labeling. (a) pole-like object labeling based on linear features, (b) building and vegetation labeling based on roughness, and (c) building and vegetation labeling after voxelization (orange and yellow points are changed vegetation and building points, respectively).}
\end{figure*}

\section{Methodology}
\label{sec:method}

\subsection{Overview}
\label{sec:method_overview}

This paper proposes an urban GeoBIM construction strategy that integrates LiDAR point clouds and as-designed BIM models, which involves instance segmentation of urban objects and matching of LiDAR data and BIM models (Fig.~\ref{fig:overview}). The focus of this construction strategy is on buildings and pole-like objects, considering the need for the management of fixed infrastructure equipment. In the instance segmentation process, an automated labeling strategy of training samples is first adopted to detect objects from LiDAR point clouds. Then, a weakly supervised framework is used for semantic segmentation, individual objects, meanwhile, are segmented based on graph theory. Next, coarse alignment and fine registration processes are employed to match LiDAR point clouds and as-designed BIM models of pole-like objects and buildings.

\subsection{Instance segmentation of LiDAR point cloud}
\label{sec:method_pcis}

To detect individual objects from urban point clouds, this paper implements an instance segmentation process that combines semantic segmentation and point cloud clustering, in which urban objects include unclassified (others), buildings, vegetation, ground, and pole-like (including traffic signal poles and streetlight) categories.

\subsubsection{Training sample labeling}
\label{sec:method_sample}

As a supervised classification strategy is adopted in this study, the labeling of training samples is a primary challenge to be addressed. Manually labeling training samples in urban scenes can be laborious and time-consuming due to the complexity of the scenes. In practice, after filtering out the ground points, some objects in urban scenes may be isolated and not connected to others. Additionally, different objects typically exhibit unique geometric properties, for example, pole-like objects can be described by lines, buildings can be described by planes, and vegetation points show discrete characteristics. Thus, an adaptive automated labeling method based on geometric attributes is presented in this paper. Specifically, ground points are detected using the cloth simulation filter algorithm \citep{rs8060501}, and then the non-ground points are clustered using the connected component labeling algorithm (Fig.~\ref{fig:ground_filter}). These steps are implemented in the software CloudCompare, where both cloth resolution and classification threshold are set to 0.5 m, and both Octree level of connected components and the minimum points per component are set to 10.

For pole-like objects, line features are fitted to determine streetlights and traffic signal poles. Typically, the height of a streetlight is greater than that of a traffic signal pole. Additionally, the bodies of streetlights and traffic signal poles can be described as lines, and the projection of a streetlight on the horizontal plane can also be described as a line (Fig.~\ref{fig:pole_sample}). Thus, when the height of a cluster is greater than 5 m, a line is fitted for points with elevation differences greater than 1 m from the highest point, and the percentage of points within 1 m from the fitted line is calculated. All points of the cluster, meanwhile, are projected onto the horizontal plane and fitted with a line, and the percentage of points within 0.5 m from the fitted line is calculated. If both percentages are greater than 0.9, then the cluster is considered a streetlight. Similarly, when the height of a cluster falls within the range of 3 m to 4 m, a line is fitted for points with elevation differences greater than 2 m from the highest point in the cluster, and if the percentage of points within 0.2 m from the fitted line is greater than 0.9, then the cluster is considered as a traffic signal pole.

\begin{figure*}[b!]
\centering
\includegraphics[width=0.95\linewidth]{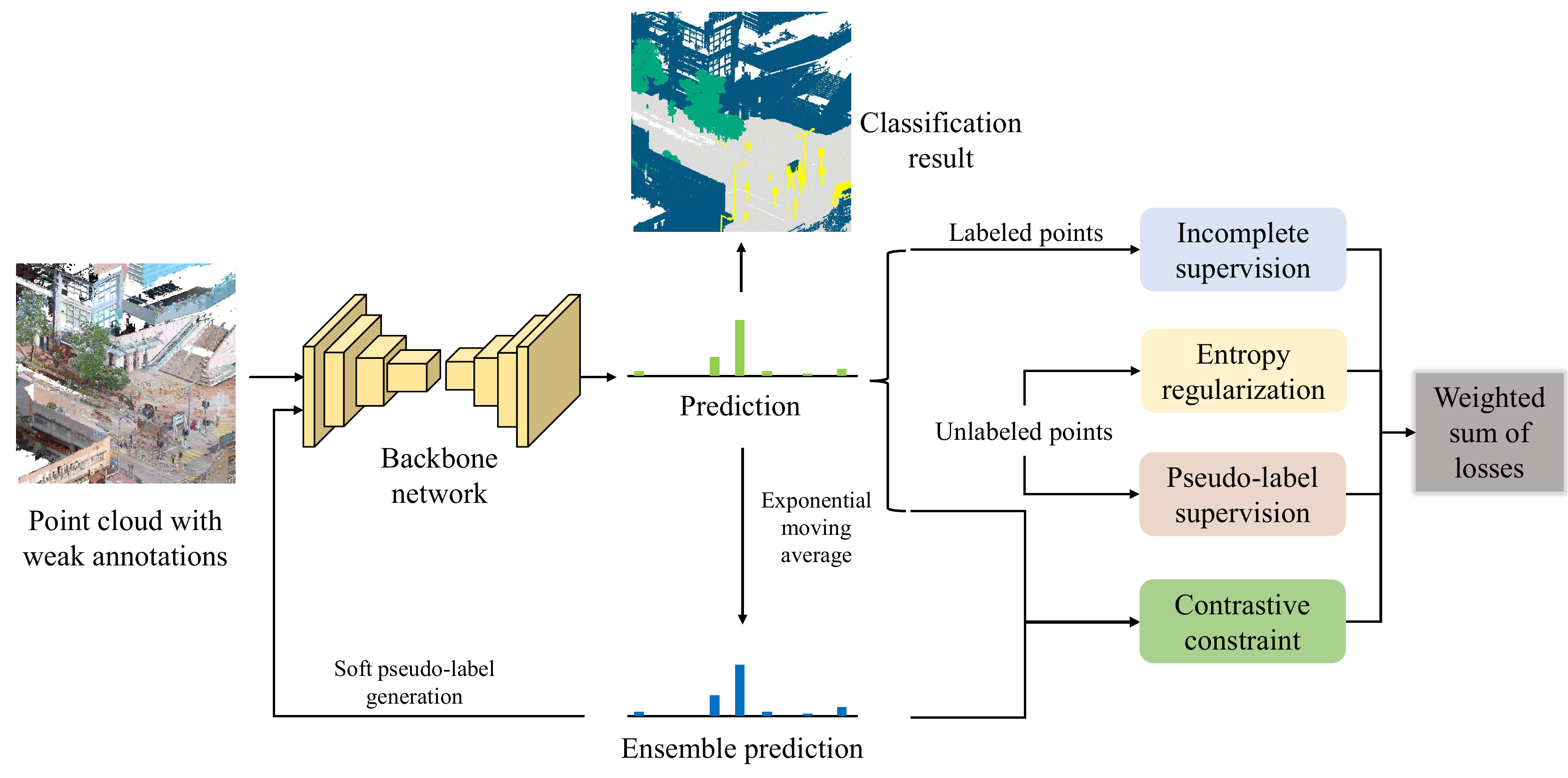}
\caption{Pipeline of the weakly supervised method, including entropy regularization, contrastive constraint, and online soft pseudo-labeling.}
\label{fig:wsss}
\end{figure*}

The local surface of the building is typically smooth, while that of vegetation is rough. Therefore, surface roughness is used as a criterion for determining buildings and vegetation. Furthermore, due to the connectivity of these clusters, they often have large sizes and point amounts. For a point $p_i\left(x_i,y_i,z_i\right)$ in these clusters, the point roughness $r_i$ can be calculated based on the eigenvalues of a covariance matrix $C_{3\times3}$, where $C$ is defined as:

\begin{equation}\label{equ:L_cov}
C=\frac{1}{n}\sum_{i=1}^{n}{\left(p_i-\bar{p}\right)\left(p_i-\bar{p}\right)}^T
\end{equation}
where $\bar{p}=\left(\frac{1}{n}\right)\sum_{i=1}^{n}p_i$ is the average of n neighboring points of $p_i$, and then $r_i$ is defined as:

\begin{equation}\label{equ:rough}
r_i=\frac{\lambda_2}{\lambda_0+\lambda_1+\lambda_2}
\end{equation}
where $\left(\lambda_0,\lambda_1,\lambda_2\right)$ are eigenvalues of C sorted in descending order as $\lambda_0\geq\lambda_1\geq\lambda_2$. To ensure accurate labeling, strict thresholds are set to determine semantic information, if the roughness $r_i$ of a point is less than 0.1, it will be considered a building point, and if the roughness $r_i$ is greater than 0.15, the point will be considered as a vegetation point.

Based on the strict roughness threshold, building, and vegetation points can generally be detected from clusters with large sizes and point amounts. However, this approach may result in a loss of detailed information and underfitting of deep learning networks due to a lack of data. (Fig.~\ref{fig:build&veg_label}). To address this, the paper further labels objects by voxelization. The raw point cloud is voxelized with a size of 2 m, and when there is only one labeled class in a voxel and its adjacent voxels have no labeled points or only points of the same class, then all points in the voxel are labeled as that class (Fig.~\ref{fig:build&veg_voxel_label}). Moreover, there may be other objects in the scene, such as vehicles and humans, that are not part of the infrastructure and are structurally complex. In such cases, these objects are manually labeled as a separate class, i.e., others.

\subsubsection{Weakly supervised semantic segmentation}
\label{sec:method_wsss}

Given that reliable labels are available for only a fraction of points, we incorporate a weakly supervised framework with a deep learning-based neural network to segment point clouds as described in our previous work \citep{WANG2022237}, which is illustrated in Fig.~\ref{fig:wsss}. In this framework, KPConv \citep{9010002} is utilized as the backbone network. As in fully supervised learning, we first apply cross-entropy loss calculation and backpropagation to weakly labeled points. To employ latent information from unlabeled data, three weakly supervised constraints are adopted to learn accurate point cloud semantics. 

Specifically, entropy regularization \citep{10.5555/2976040.2976107} is first incorporated to minimize the uncertainty of the posterior probability by reducing the Shannon entropy $H$ of point-wise predictions. This is based on the finding that unlabeled data contains less information with high uncertainty when prediction classes overlap. Furthermore, inspired by contrastive learning, we introduce consistency constraints to produce stable predictions by enforcing consistency between the current semantic prediction and its ensemble value. Online soft pseudo-labeling is used to expand the labeled pool in an efficient and nonparametric way. Pseudo labels are defined as the category with the maximum prediction probability of ensemble predictions. Instead of empirically setting a fixed threshold for pseudo-label selection, we leverage all unlabeled points and soften their pseudo-labels with reliability weights that are directly inferred from prediction entropy. All the proposed losses jointly participate in backpropagation during training, enhancing model performance without sacrificing efficiency.

\subsubsection{Graph-based point cloud clustering}
\label{method:cluster}

Different objects can be distinguished based on the semantic segmentation process, but obtaining individual objects still requires the instance segmentation process. In urban scenes, buildings and pole-like objects are often isolated, but individual object points are continuously distributed. Therefore, a graph-based clustering strategy is used to segment individual objects and optimize semantic segmentation results. The conventional graph-based clustering approach typically segments the point cloud based on distance information, which performs well for isolated objects but struggles with connected objects. To enhance applicability, a combination of topological and geometric relationships between points is proposed for graph-based clustering, where the topological relationship is described by distance, and the geometric relationship is described by roughness and anisotropy. The connected component labeling algorithm is first used to cluster each semantic class, and the resulting clusters are used as primitives. Subsequently, an undirected graph is constructed based on these clusters, with each cluster represented as a vertex and connections between vertices referred to as edges. The minimum distance, average roughness, and anisotropy of each cluster are calculated. The distance is calculated as the Euclidean distance, roughness is defined as Eq.~\ref{equ:rough}, and anisotropy is defined as $a_i=\left(\lambda_0-\lambda_2\right)/\lambda_0$. The connection weight between each pair of vertices can be calculated by:

\begin{equation}\label{equ:connection_w}
    w_{ij}=e^{-\frac{1}{3}\left(\left(\frac{d_{ij}}{\sigma_1}\right)^2+\left(\frac{r_i-r_j}{\sigma_2}\right)^2+\left(\frac{a_i-a_j}{\sigma_3}\right)^2\right)}
\end{equation}
where $w_{ij}$ and $d_{ij}$ represent the connection weight and the minimum distance between vertexes $i$ and $j$, respectively, $\left(\sigma_1,\sigma_2,\sigma_3\right)$ represent the standard deviations of distance, roughness, and anisotropy. If $w_{ij}$ is greater than a threshold (0.75 in this paper), the adjacent vertices will be considered as belonging to the same cluster. In practice, the class of the cluster is defined by the class of the vertex with the largest size.

\begin{figure*}[t!]
\centering
\begin{subfigure}[b]{0.49\textwidth}
    \centering
    \includegraphics[width=1\linewidth]{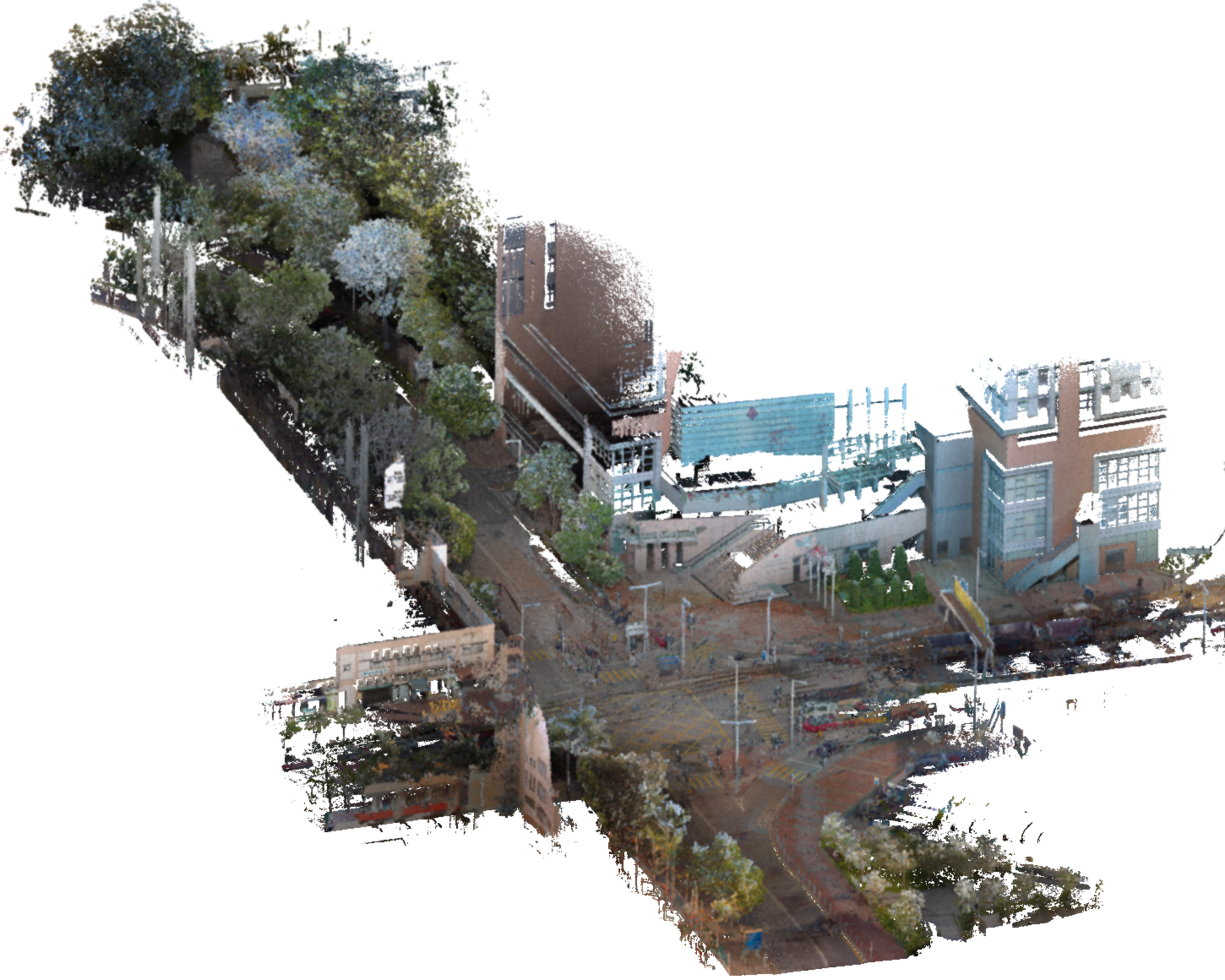}
    \caption{}
    \label{fig:hk_data}
\end{subfigure}
\hfill
\begin{subfigure}[b]{0.49\textwidth}
    \centering
    \includegraphics[width=1\linewidth]{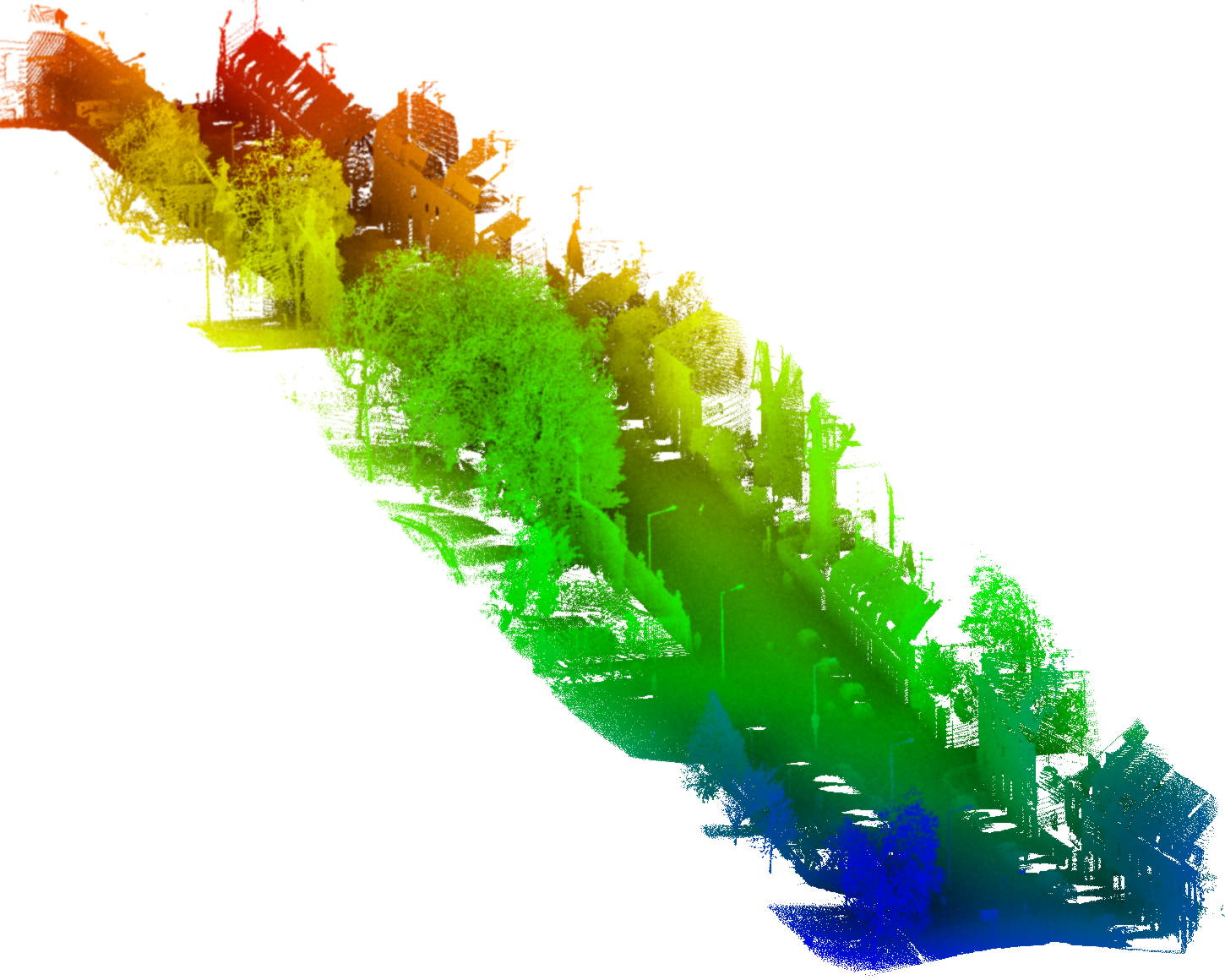}
    \caption{}
    \label{fig:lille}
\end{subfigure}
\caption{The two test datasets, (a) shows the colored LiDAR point cloud in Dataset 1, and (b) shows Dataset 2.}
\label{fig:data}
\end{figure*}

\subsection{Matching of LiDAR point clouds and BIM models}
\label{method:match}

Due to the difference in structure, the matching of LiDAR point clouds and BIM models for different objects is performed using different methods. Aligning building LiDAR point clouds and BIM models is more challenging due to the complex structures of buildings compared to pole-like objects. Therefore, a robust coarse alignment method for LiDAR point clouds and as-designed BIM models of buildings is proposed, inspired by the 4PCS algorithm. To increase efficiency, building edges are first detected using the Angle Criterion approach \citep{Bendels2006DetectingHI} based on the open-source Point Cloud Library (PCL). Keypoints are then determined by calculating the angles between a point and its neighboring points, which replace the dense raw point cloud. The matching of LiDAR keypoints and BIM models is performed, with the LiDAR data set as the reference and BIM models as the aligned data. Specifically, corresponding 4-point pairs that are not co-planar are detected from the LiDAR and BIM point clouds. The distances between all points of the BIM model and their closest neighbors in the LiDAR data are counted, and 4-point pairs that obtain the smallest distance are presented as the optimal correspondence. The pseudo-code of the alignment process is shown in Algorithm~\ref{alg}.

\begin{algorithm}[t!]
\small
\caption{Coarse alignment of LiDAR point cloud and BIM model}\label{alg}
\KwIn{
\\ $\ast$ \textit{Key points in LiDAR data $\bm{L}$ and as-designed BIM point cloud $\bm{B}$}
\\ $\ast$ \textit{Angle range $\bm{\theta}$ and distance threshold $\eta$}
}
\KwOut{\textit{Rotation and translation matrices} $\left[\textbf{\textit{R}}\in\mathbb{R}^{3\times3},T\right]$}
\tcc{Producing 2-point sets $\bm{LL}$ and $\bm{BB}$ by distances between points}
\For{$i\gets0$ \KwTo $length(\bm{L})(\bm{B})-1$}{
    \For{$j\gets i+1$ \KwTo $length(\bm{L})(\bm{B})$}{
        $d_{ij}=\left|\bm{L}_i-\bm{L}_j\right| (\left|\bm{B}_i-\bm{B}_j\right|)$ \\
        \If{$d_{ij}>\eta$}{
            $\bm{LL}\gets\left(\bm{L}_i,\bm{L}_j\right)$, $ \bm{BB}\gets\left(\bm{B}_i,\bm{B}_j\right)$
            }
        }     
    }  
\tcc{Generating 4-point sets in LiDAR $\bm{LS}$ and BIM $\bm{BS}$}
\For{$i\gets0$ \KwTo $length(\bm{LL})(\bm{BB})-1$}{
   \For{$j\gets i+1$ \KwTo $length(\bm{LL})(\bm{BB})$}{
        \textit{Distance $d$ and angle $\alpha$ between 2-point pairs} \\
        \textit{Coplanarity of 2-point pairs} \\
        \If{$d>\eta/2$ $\&\&$ $\alpha\in\bm{\theta}$ $\&\&$ non-coplanar}{
            $\bm{LS}\gets\left(\bm{LL}_i,\bm{LL}_j\right)$, $ \bm{BS}\gets\left(\bm{BB}_i,\bm{BB}_j\right)$
                 }
        }   
    }
\tcc{Determining corresponding 4-point pairs $\bm{PP}$ in LiDAR and BIM}
\For{$i\gets0$ \KwTo $length(\bm{BS})$}{
    \For{$j\gets0$ \KwTo $length(\bm{LS})$}{
        $\alpha_{ij}=\left|\alpha_{Bi}-\alpha_{Lj}\right|$, $d_{ij}=\left|d_{Bi}-d_{Lj}\right|$ \\
        \If{$\alpha_{ij}<1$ $\&\&$ $d_{ij}<0.5$}{
            $\bm{PP}\gets\left({\bm{BS}}_i,{\bm{LS}}_j\right)$
        }
    }
}
\tcc{Calculating $\left[\bm{R},T\right]$ of LiDAR point cloud and BIM models}
\For{$i\gets0$ \KwTo $length(\bm{PP})$}{
    ${\bm{LS}}_i\left(\in\bm{PP}\right)=\bm{R}\cdot{\bm{BS}}_i\left(\in\bm{PP}\right)+T$ \\
    $d_{LB}=\left|\bm{LS}-T\left(\bm{BS}\right)\right|$ \\
    \tcc{The optimal coarse alignment of LiDAR and BIM data}
    \If{$d_{LB}$ is min}{
        return $\left[\bm{R},T\right]$
    }
}

\end{algorithm}

Accurate matching results are generally difficult to obtain by the coarse alignment process and a fine registration process becomes necessary. Conventional fine registration algorithms like ICP may struggle to achieve accurate registration due to sparse BIM point clouds and incomplete LiDAR data. Therefore, in this study, a further fine registration process based on the point-to-plane association strategy is proposed, building upon the ICP registration results. For a point P in BIM models, three points $\left\{Q_1,Q_2,Q_3\right\}$ from the raw LiDAR data that are closest to the point P and not on the same line are searched for, and the point-to-plane distance $d$ between P and the plane formed by $\left\{Q_1,Q_2,Q_3\right\}$ can be computed by

\begin{equation}\label{equ:p2p_dis}
    d=\frac{\left|\left(P-Q_1\right)\times\left(\left(P-Q_2\right)\times\left(P-Q_3\right)\right)\right|}{\left|\left(P-Q_2\right)\times\left(P-Q_3\right)\right|}
\end{equation}
where the matching is accurate, $d$ will be small. Additionally, the transformation relationship between LiDAR point clouds $\textbf{\textit{X}}_{Lidar}$ and BIM models $\textbf{\textit{X}}_{BIM}$ can be defined as

\begin{equation}\label{equ:transform_relation}
    \textbf{\textit{X}}_{Lidar}=\textbf{\textit{R}} \cdot \textbf{\textit{X}}_{BIM}+T
\end{equation}
where $\textbf{\textit{R}}$ and $T$ are unknowns and represent rotation and translation, which can be solved by minimizing error $e$ of Eq.~\ref{equ:p2p_dis}. Then, the Levenberg-Marquardt (L-M) algorithm \citep{9118969} is used to calculate these unknown parameters.

\begin{equation}\label{equ:error}
    e=\arg \min_{e} \frac{1}{2} \sum \left\|d(\textbf{\textit{R}}, T)-0\right\|^2
\end{equation}

Once $\left[\textbf{\textit{R}},T\right]$ is calculated, BIM models can be obtained the geo-referenced information. For each pole-like object, the center of mass is calculated and set as the origin. Then, three mutually perpendicular eigenvectors are computed using the principal component analysis, which serves as the three main directions of each point cloud. Subsequently, the LiDAR data and BIM model are coarsely aligned based on the calculated origin and main directions. Due to the simple structures, the ICP algorithm is directly employed for the fine registration of LiDAR data and BIM models of pole-like objects.

\subsection{Performance evaluation criteria}
\label{sec:method_val}

The accuracy evaluation criteria for semantic segmentation results employed in this work involved the overall accuracy (OA) and F1 scores. OA indicates the percentage of predictions correctly classified, and the F1 score stands for the harmonic mean of the precision ($p$) and recall ($r$), formulated as:

\begin{equation}\label{equ:val}
\begin{aligned}
& p=\frac{TP}{TP+FP}\\
& r=\frac{TP}{TP+FP}\\
& F1=2\times\frac{p\times r}{p+r}\\
\end{aligned}
\end{equation}
where $TP$, $FP$, and $FN$ are true positives, false positives, and false negatives, respectively. For the evaluation of urban GeoBIM construction and matching of LiDAR data and BIM models, the distance between corresponding features is counted, including the minimum, maximum, average, and root mean square error ($RMSE$).

\begin{equation}\label{equ:rmse}
    RMSE=\sqrt{\frac{\sum_{i=1}^{n}d_i^2}{n}}
\end{equation}
where $n$ is the number of objects and $d_i$ is the distance between correspondences.

\begin{figure}[b!]
\centering
\begin{subfigure}[b]{0.49\linewidth}
    \centering
    \includegraphics[width=1\linewidth]{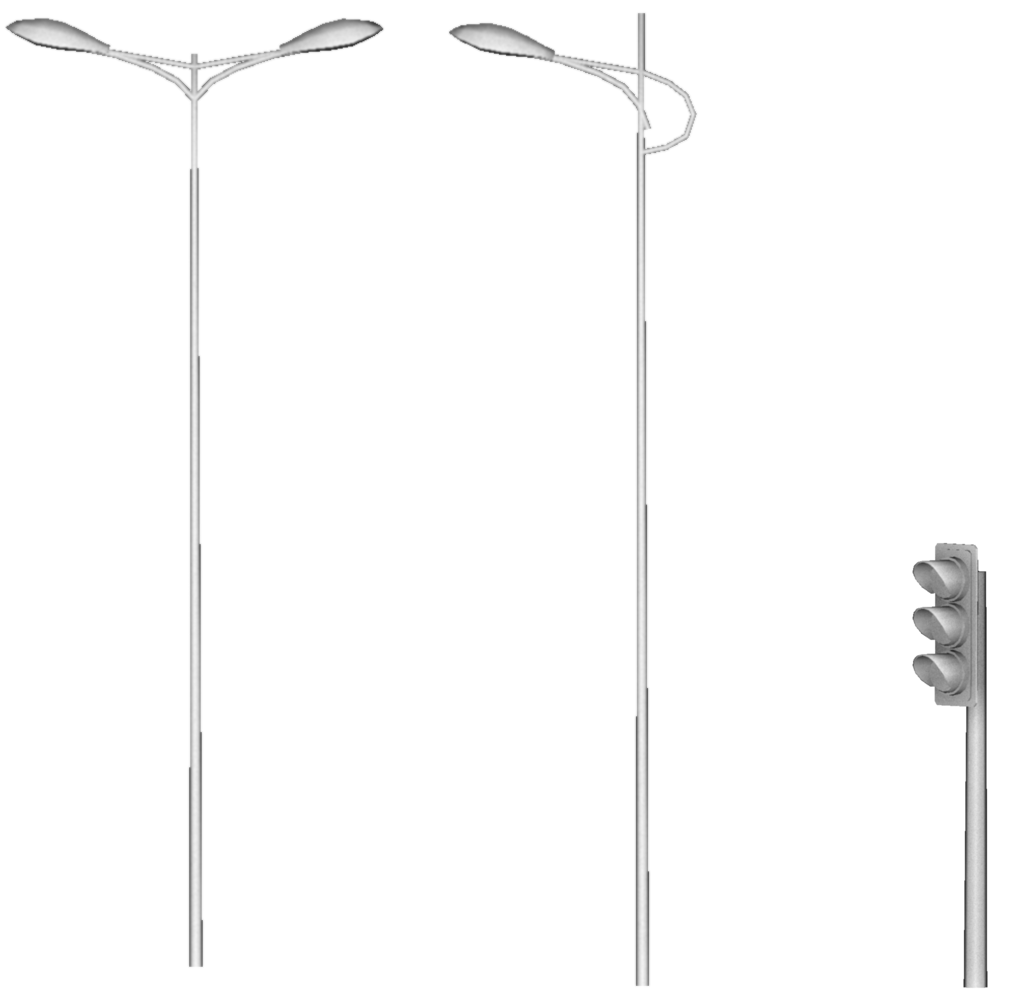}
    \caption{}
    \label{fig:pole_bim}
\end{subfigure}
\hfill
\begin{subfigure}[b]{0.49\linewidth}
    \centering
    \includegraphics[width=1\linewidth]{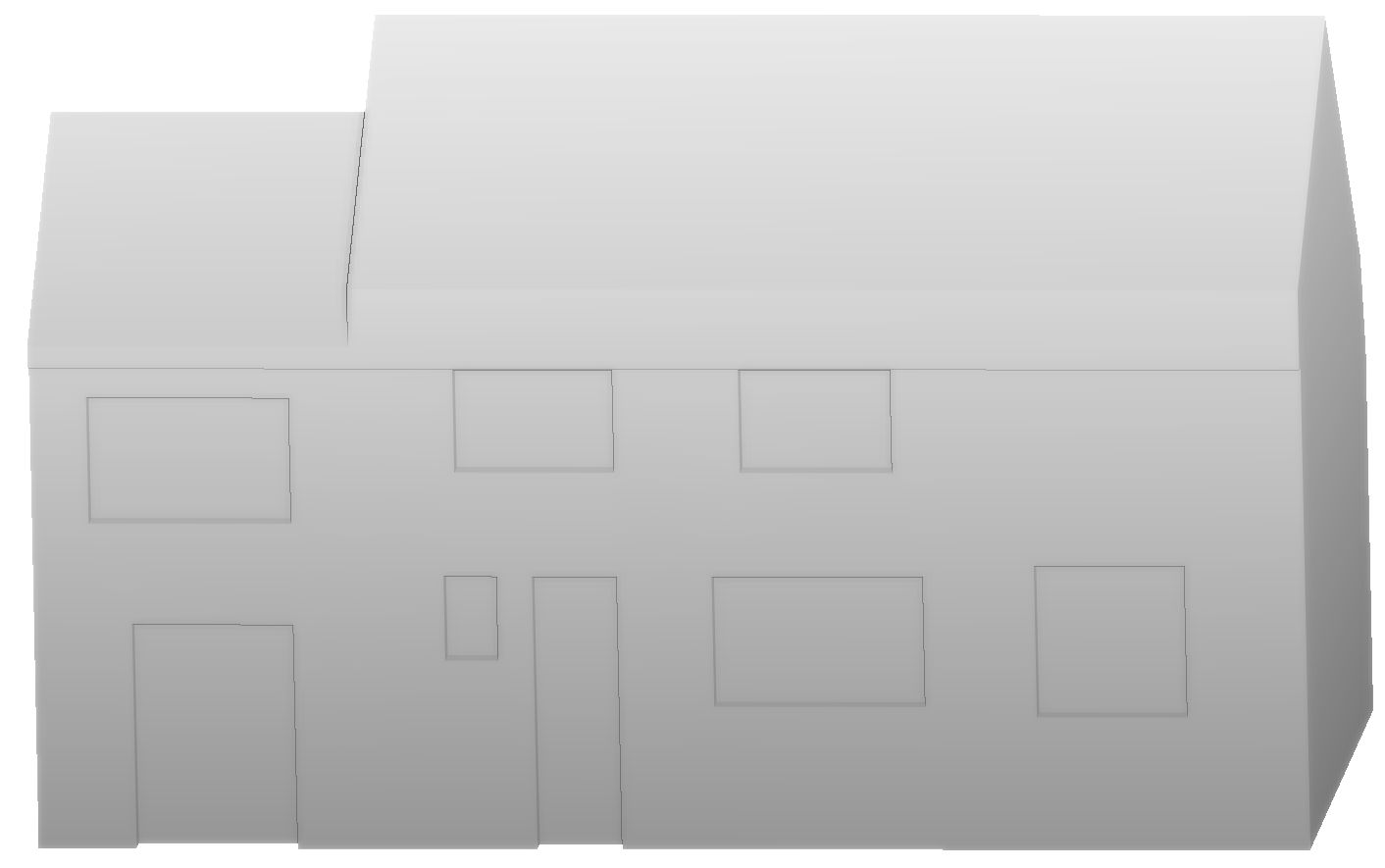}
    \caption{}
    \label{fig:build_bim}
\end{subfigure}
\caption{As-designed BIM models. (a) shows pole-like objects, including streetlights and a traffic signal pole, (b) shows a building BIM model.}
\label{fig:bim_model}
\end{figure}

\begin{figure*}[b!]
\centering
\begin{subfigure}[b]{1.0\linewidth}
    \centering
    \includegraphics[width=0.7\linewidth]{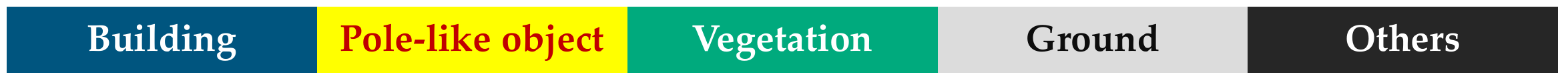}
\end{subfigure}
\hfill
\begin{subfigure}[b]{0.49\linewidth}
    \centering
    \includegraphics[width=1\linewidth]{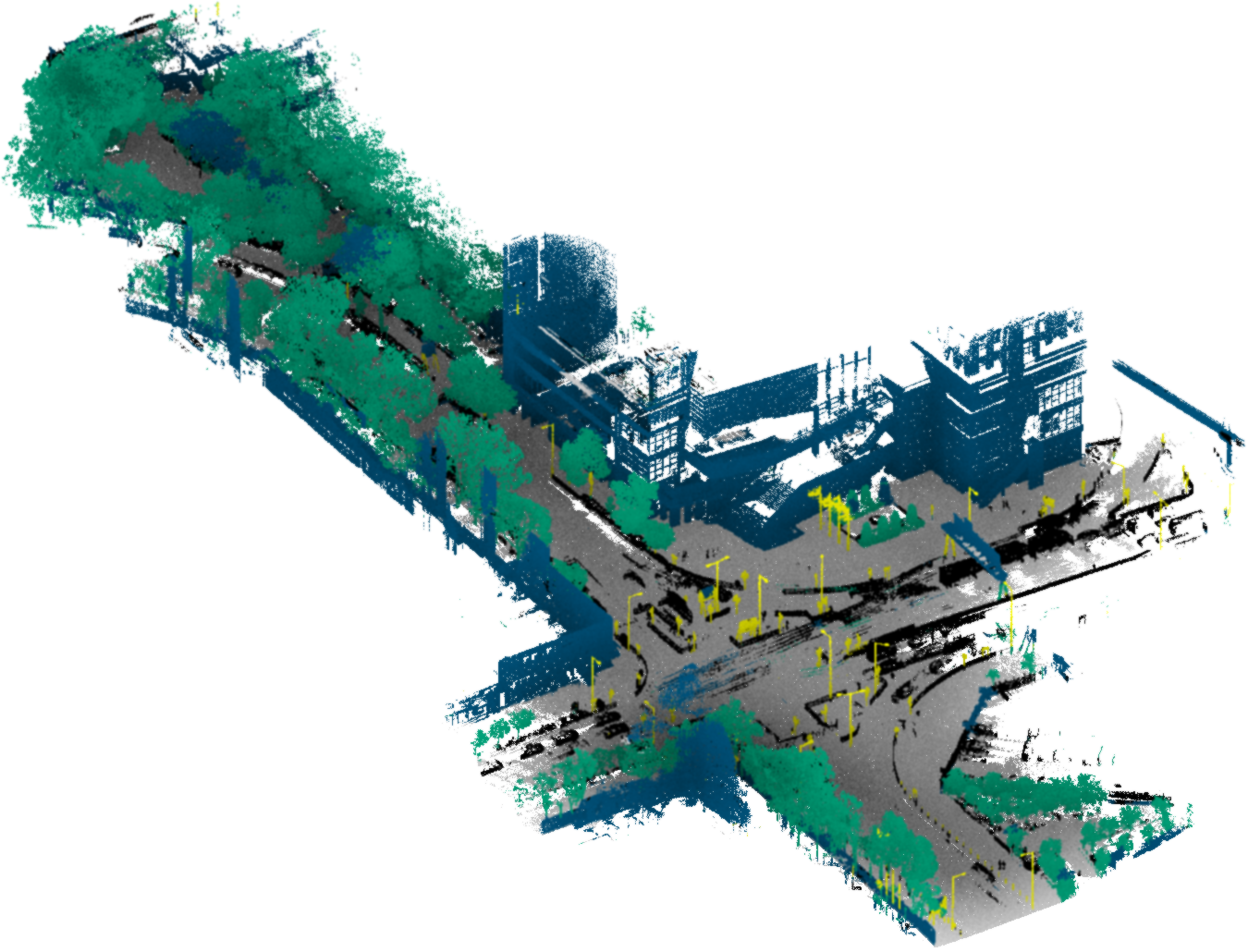}
    \caption{}
\end{subfigure}
\hfill
\begin{subfigure}[b]{0.49\linewidth}
    \centering
    \includegraphics[width=1\linewidth]{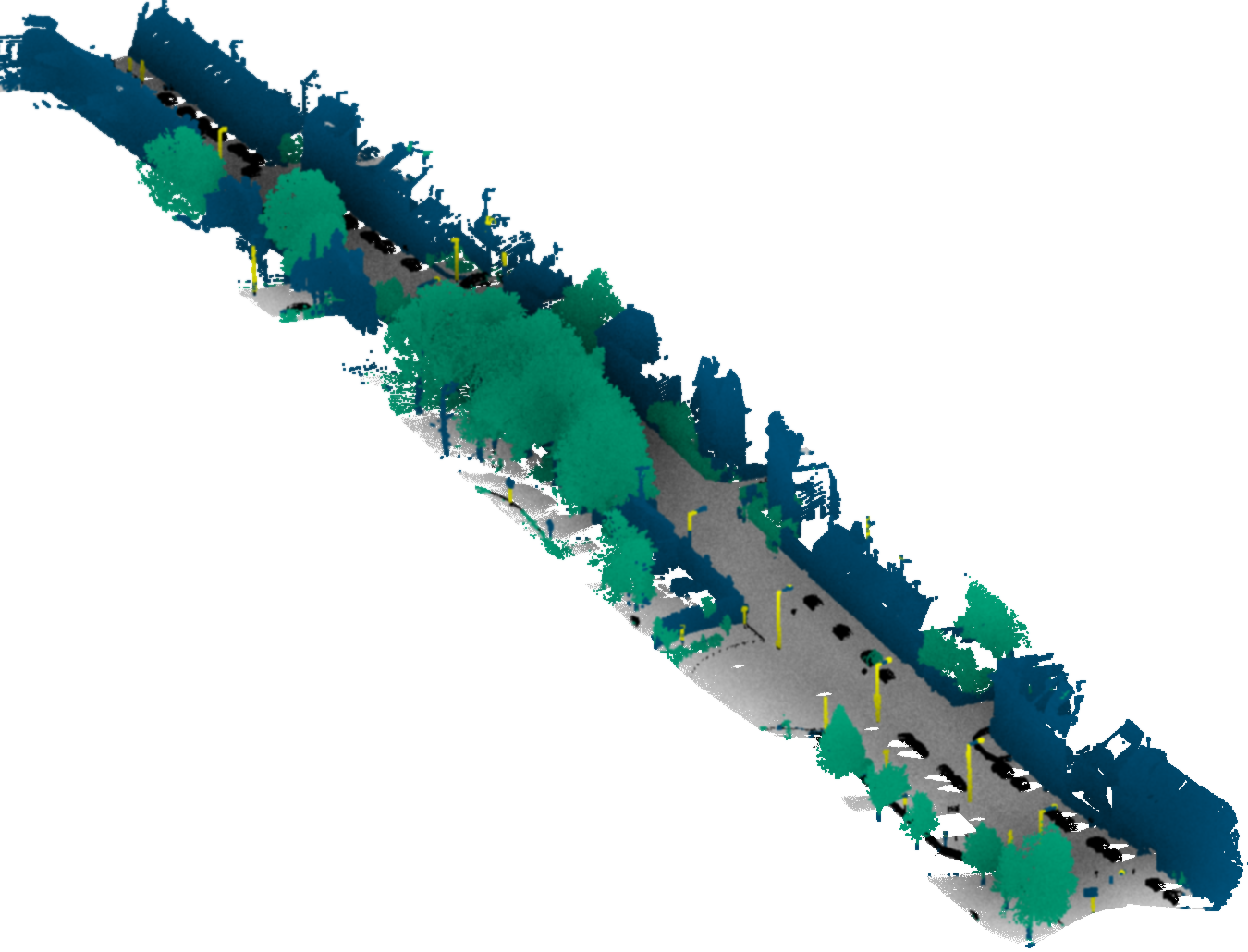}
    \caption{}
\end{subfigure}
\caption{Semantic classification results of Dataset 1 (a) and Dataset 2 (b).}
\label{fig:ss_res}
\end{figure*}

\begin{table*}[b!]
\footnotesize
\caption{The Confusion matrix of classification results with per-class precision, recall, and F1 score (\%).}
\label{tab:ss_res}
\begin{tabularx}{\textwidth}{XXXXXXXXX}
\hline
\multirow{2}{*}{Dataset} & \multirow{2}{*}{Criteria (\%)} & \multicolumn{5}{c}{Classes}                     & \multirow{2}{*}{Avg. F1 (\%)} & \multirow{2}{*}{OA (\%)} \\ \cline{3-7}
                         &                                & Others & Ground & Building & Vegetation & Pole-like  &                               &                          \\ \hline
\multirow{3}{*}{1}       & Precision                      & 99.08  & 97.25  & 71.15    & 96.68      & 95.30 & \multirow{3}{*}{85.00}        & \multirow{3}{*}{90.48}   \\
                         & Recall                         & 39.81  & 99.77  & 99.70    & 96.28      & 85.62 &                               &                          \\
                         & F1 score                       & 56.80  & 98.49  & 83.04    & 96.48      & 90.20 &                               &                          \\ \hline
\multirow{3}{*}{2}       & Precision                      & 87.61  & 99.08  & 80.83    & 89.20      & 98.04 & \multirow{3}{*}{77.55}        & \multirow{3}{*}{91.24}   \\
                         & Recall                         & 37.49  & 99.29  & 98.39    & 90.71      & 40.53 &                               &                          \\
                         & F1 score                       & 52.51  & 99.18  & 88.75    & 89.95      & 57.35 &                               &                          \\ \hline
\end{tabularx}
\end{table*}

\section{Results}
\label{sec:res}

\subsection{Data preparation}
\label{sec:res_data}

Two datasets, namely Dataset 1 and Dataset 2, from different urban locations, are selected for evaluating the effectiveness of this method. Dataset 1 was acquired from Hong Kong, China, while Dataset 2 was obtained from Lille, France. In each dataset, geo-referenced LiDAR data were collected using mobile laser scanning systems, and BIM models were created manually. In this study, the fixed infrastructures in both locations mainly consist of buildings and pole-like objects, including traffic signal poles and streetlights.

\subsubsection{LiDAR point clouds}
\label{sec:res_pc}

In Dataset 1, the LiDAR point cloud was acquired using a backpack LiDAR system that integrates an inertial measurement unit (IMU), GNSS, and a laser scanner (VELODYNE VLP-16). The system provides centimeter-level position accuracy, with a maximum scan range of up to 100 m, a field of view of 360\degree × 30\degree, and angle resolutions of 0.2\degree in the horizontal direction and 2\degree in the vertical direction. The scene area of Dataset 1 is 235 m × 225 m, and the point cloud resolution is approximately 0.01 m. However, due to the horizontal placement of the laser scanner, acquiring superstructure information from the backpack LiDAR system in Dataset 1 is challenging (Fig.~\ref{fig:hk_data}). To evaluate the classification result, we manually annotate the data with 5 categories, including ground, building, vegetation, pole-like object, and others. Dataset 2 is obtained from the open urban point cloud dataset Paris-Lille-3D \citep{doi:10.1177/0278364918767506}, and the point cloud has a high density, with between 1,000 and 2,000 points per square meter on the ground. For this study, a street segment approximately 360 m in Lille, France, was selected for validating the proposed method (Fig.~\ref{fig:lille}). Additionally, to increase the efficiency, all LiDAR data are subsampled with a grid size of 0.03 m. Except for maintaining points of ground, buildings, natural (vegetation), and pole-like objects, we assigned the remaining points as the category of others.

\begin{figure*}[b!]
\centering
\includegraphics[width=1\linewidth]{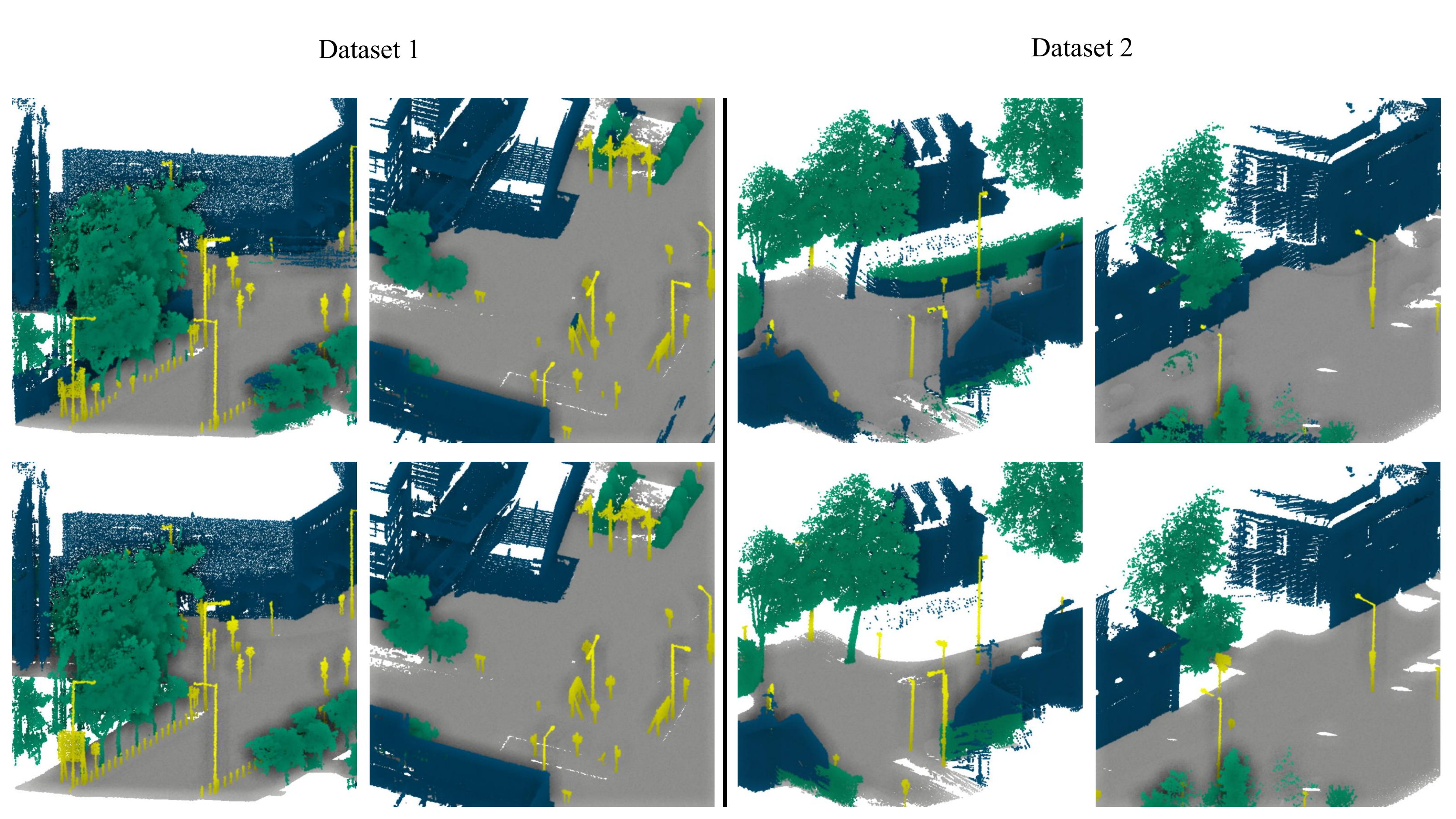}
\caption{Detailed classification results. The first row shows the predictions, while the second row presents the ground truth.}
\label{fig:ss_detail}
\end{figure*}

\begin{figure*}[b!]
\centering
\begin{subfigure}[b]{0.49\linewidth}
    \centering
    \includegraphics[width=1\linewidth]{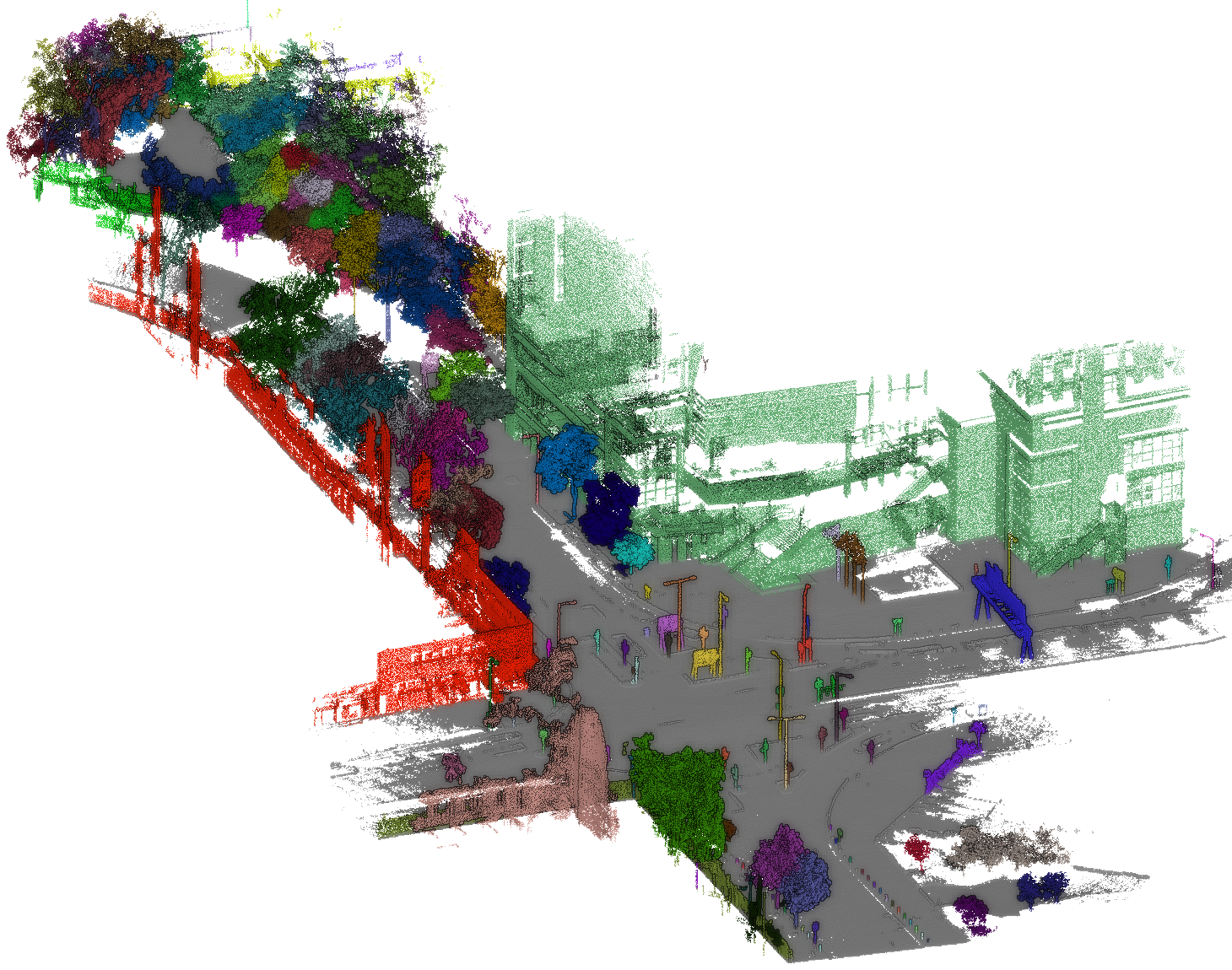}
    \caption{}
    \label{fig:is_1}
\end{subfigure}
\hfill
\begin{subfigure}[b]{0.49\linewidth}
    \centering
    \includegraphics[width=1\linewidth]{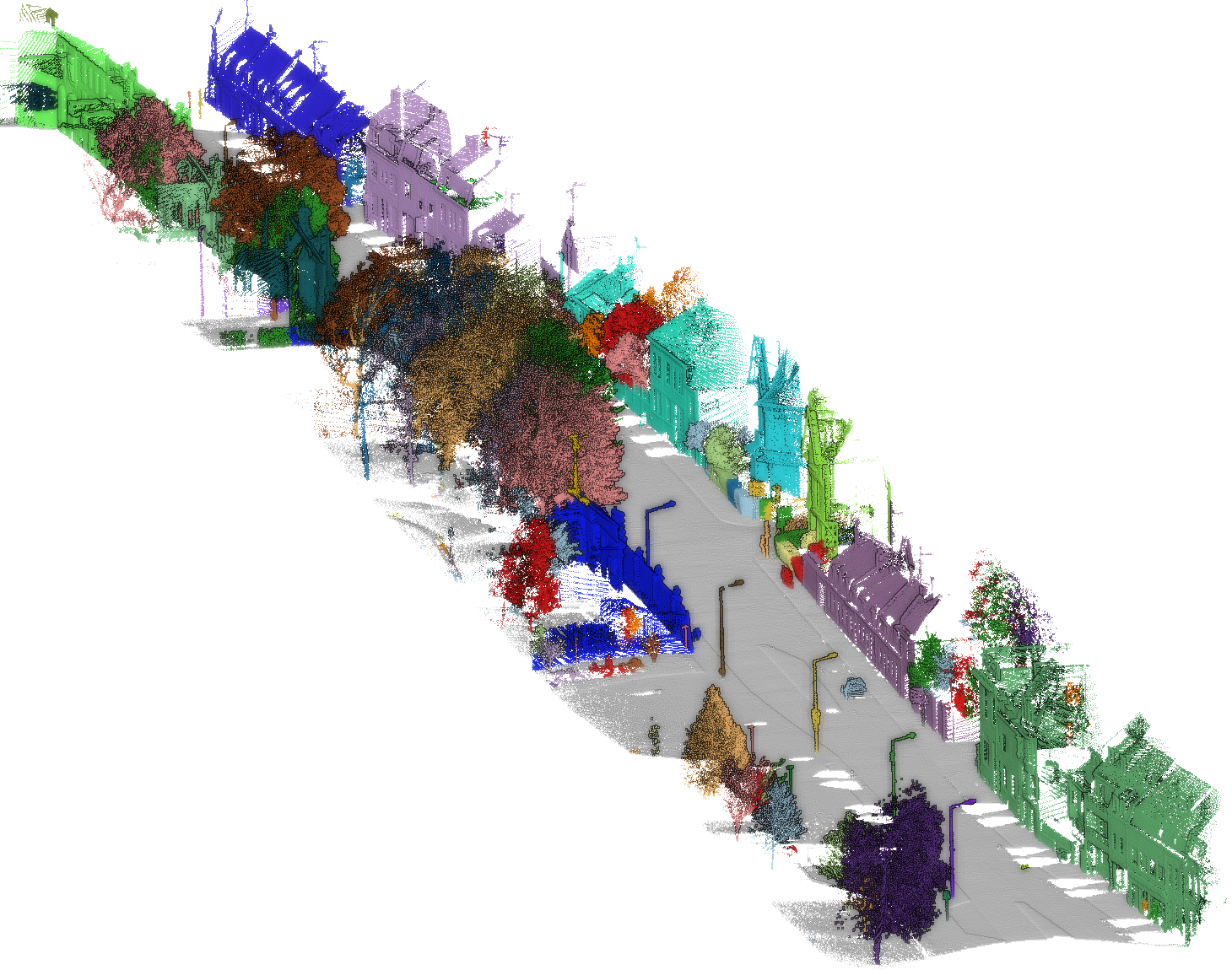}
    \caption{}
    \label{fig:is_2}
\end{subfigure}
\caption{Instance segmentation results of Dataset 1 (a) and Dataset 2 (b), different colors represent different instances.}
\label{fig:is}
\end{figure*}

\subsubsection{BIM models}
\label{sec:res_bim}

For buildings and pole-like objects, the corresponding BIM models were designed using Autodesk 3DMAX, a commercial software. However, due to the lack of as-designed BIM models, we referred to the LiDAR point cloud of each object to design coarse BIM models for verification purposes. Buildings and pole-like objects were designed as LoD2 models, taking into consideration the characteristic of LiDAR data, as shown in Fig.~\ref{fig:pole_bim} and Fig.~\ref{fig:build_bim}, respectively.

\begin{figure*}[b!]
\centering
\begin{subfigure}[b]{0.49\linewidth}
    \centering
    \includegraphics[width=1\linewidth]{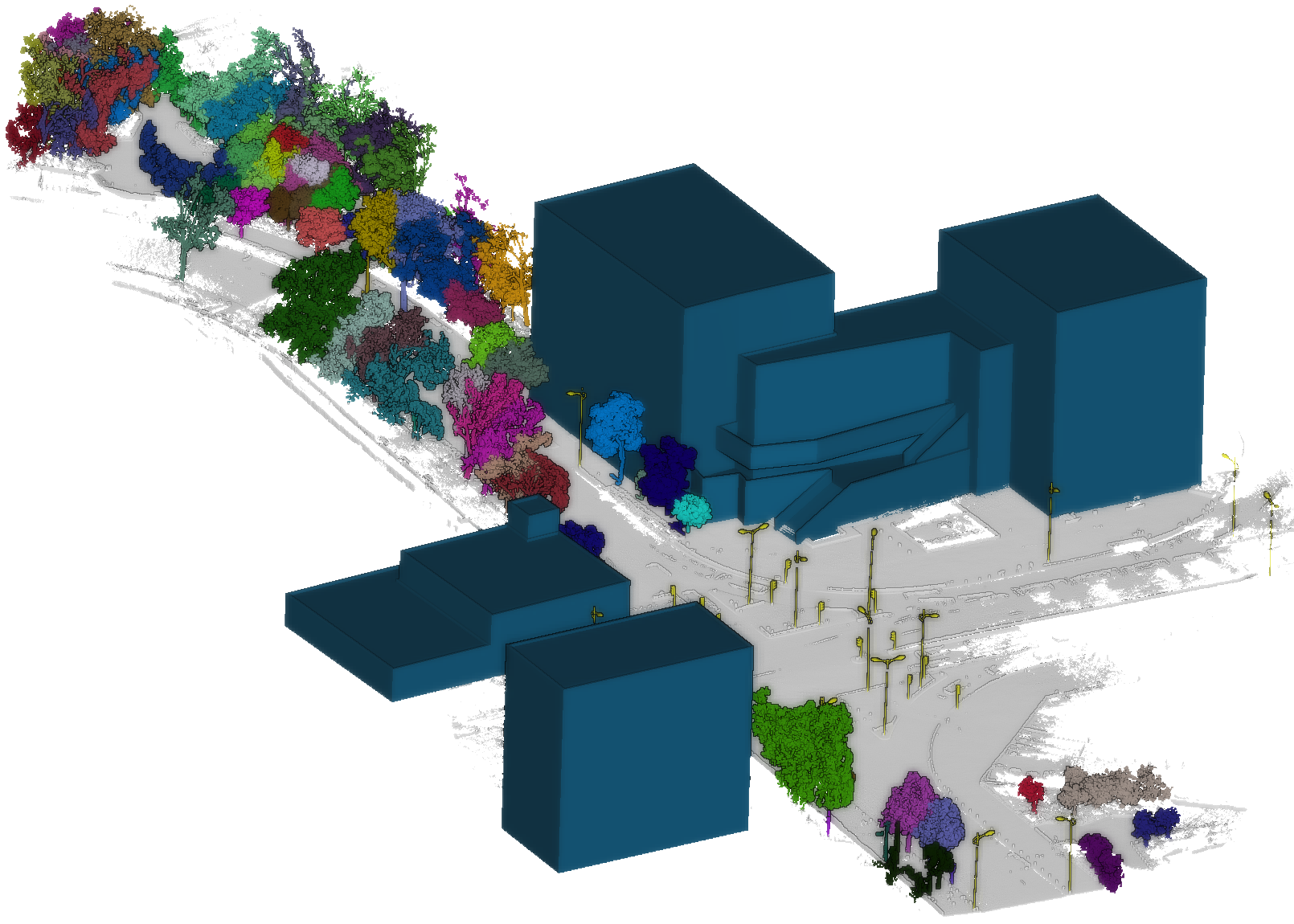}
    \caption{}
    \label{fig:bim_1}
\end{subfigure}
\hfill
\begin{subfigure}[b]{0.49\linewidth}
    \centering
    \includegraphics[width=1\linewidth]{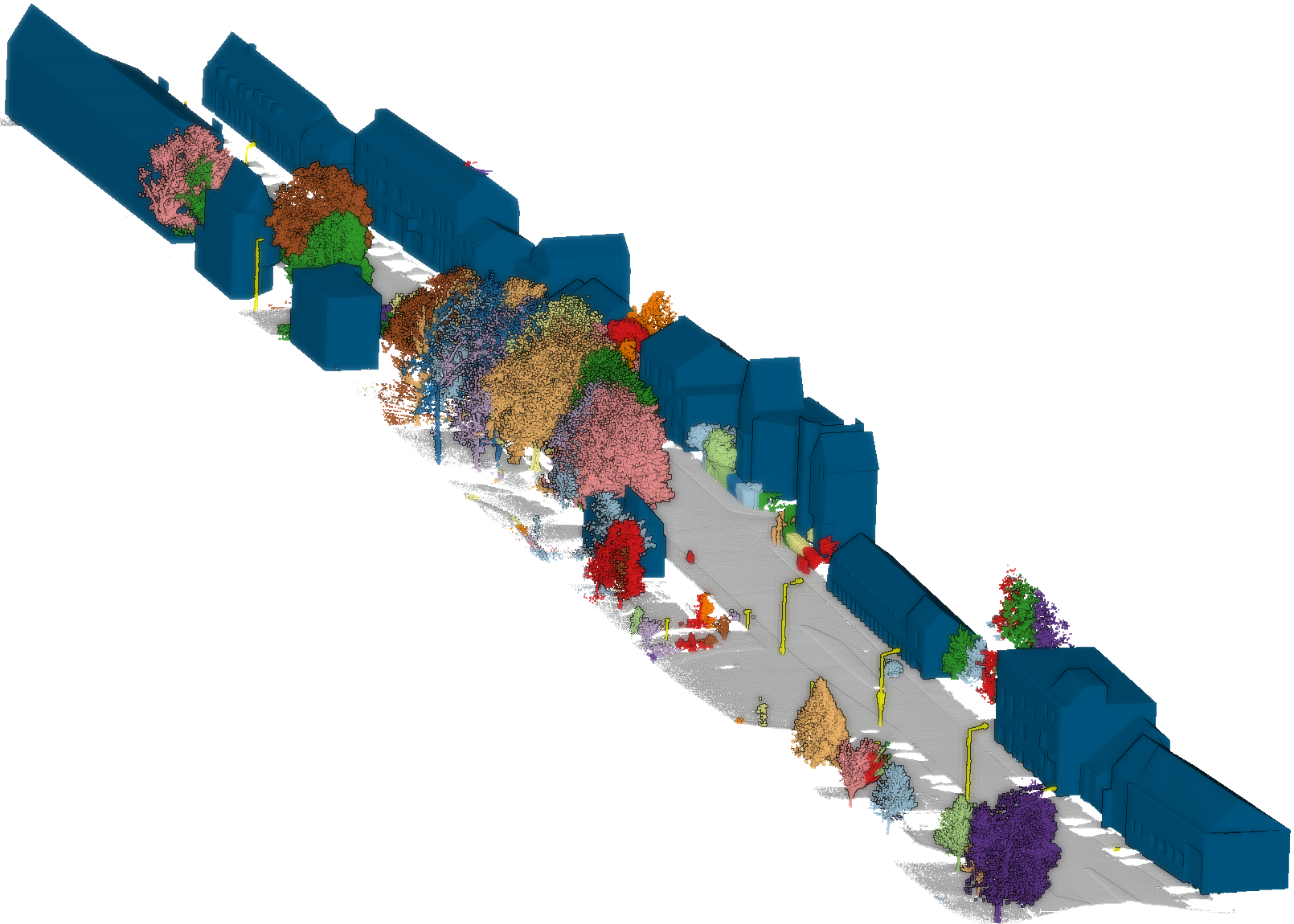}
    \caption{}
    \label{fig:bim_2}
\end{subfigure}
\caption{Urban GeoBIM construction of Dataset 1 (a) and Dataset 2 (b).}
\label{fig:geobim}
\end{figure*}

\begin{figure*}[b!]
\centering
\begin{subfigure}[b]{0.49\linewidth}
    \centering
    \includegraphics[width=1\linewidth]{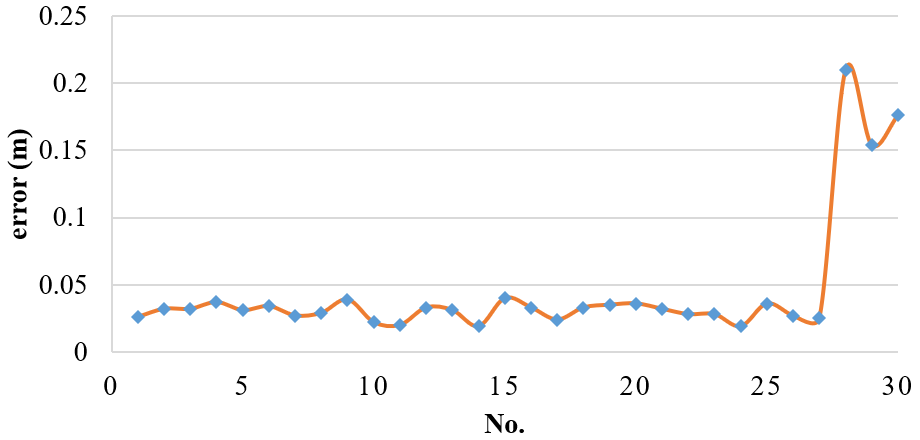}
    \caption{}
\end{subfigure}
\hfill
\begin{subfigure}[b]{0.49\linewidth}
    \centering
    \includegraphics[width=1\linewidth]{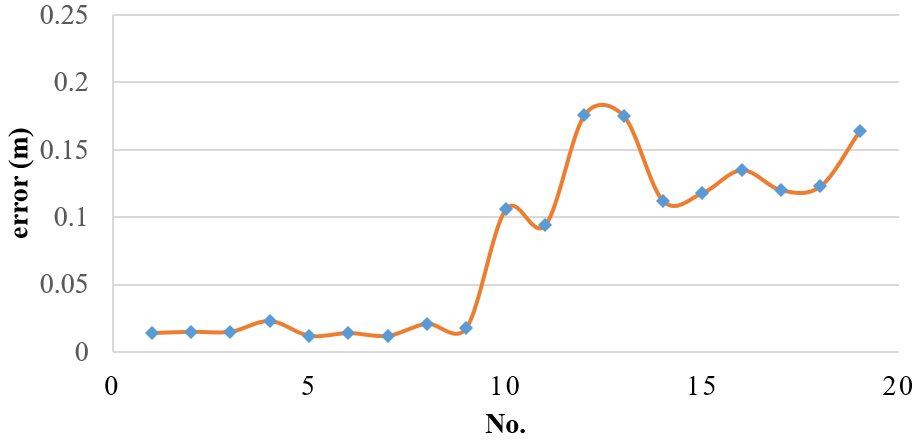}
    \caption{}
\end{subfigure}
\caption{Positioning errors of each pole-like object and building BIM model, the last three values in (a) and the last ten values in (b) are building positioning errors, and other values are positioning errors of pole-like objects, where the error is calculated based the average Euclidean distance between the corner points of BIM models and their corresponding planes in the LiDAR data.}
\label{fig:error}
\end{figure*}

\subsection{Semantic segmentation results}
\label{sec:res_ss}

We present the classification map of semantic segmentation results in Fig.~\ref{fig:ss_res}. It can be observed that the majority of points are correctly classified, indicating the effectiveness of our classification method. Detailed classification results are further illustrated in Fig.~\ref{fig:ss_detail}, where semantic boundaries are well preserved, providing effective data for instance segmentation and BIM generation. However, there are some typical errors observed in the figures. For instance, points belonging to fences, which were unclassified in this study, could be misclassified as buildings, as points associated with these two categories exhibit similar verticality. Additionally, while separated pole-like objects are well classified, those that are close to trees may be directly predicted as vegetation due to their proximity.

Furthermore, we have conducted a quantitative evaluation of classification results as shown in Table~\ref{tab:ss_res}. Categories of ground and vegetation are well classified, with high precision and recall scores. However, there are some misclassifications of unnecessary points as buildings, resulting in a relatively low precision score for the building category. This could be attributed to the limitations of MLS point cloud acquisition, where LiDAR sensors may have difficulty detecting roof areas, resulting in only the façade of buildings being preserved. As a result, some objects that were not considered in this study, such as fences, may be prone to being classified as buildings. On the other hand, due to the isolated characteristic of pole-like objects, the pole-like category shows a high precision score, although some of them may be predicted as other classes.

\subsection{Instance segmentation results}
\label{sec:res_is}

To address the challenge of weakly supervision classification, where point clouds of individual objects are not available, we propose a segment-based clustering method that combines distribution characteristics of object point clouds at local and global scales. In practice, each semantic class is segmented in turn, and segments of each class are clustered based on the proposed method. Subsequently, the primary clusters of all classes are clustered again, as illustrated in Fig.~\ref{fig:is}. This approach allows us to obtain reliable clusters of objects, despite the limitations of weak supervision, and serves as a key step toward instance segmentation and urban GeoBIM construction.

From the results, buildings, pole-like objects, and some isolated trees can be segmented effectively. Buildings and pole-like objects, being discretely distributed in most cases, can be accurately instantiated based on the spatial distance relationship. Additionally, roughness features can also aid in correctly classifying buildings and trees that are in contact with each other. However, for gathered trees, the proposed graph-based method may not be effective for segmentation. Since the vegetation category is not considered in this study, the analysis of vegetation instance segmentation is deferred for now. When it comes to pole-like objects, errors mainly arise from the semantic segmentation results. If a pole-like object is misclassified as vegetation due to its proximity to a tree, it may be difficult to optimize and instance the pole-like object based on the proposed method. Similarly, segmenting buildings that are gathered closely together can pose challenges to instance segmentation.

\begin{table*}[t!]
\footnotesize
\caption{Positioning accuracy of BIM models. ‘N’ is the number of corresponding objects.}
\label{tab:error}
\begin{tabularx}{\textwidth}{XXXXXXXXXXXX}
\hline
\multirow{2}{*}{Dataset} & \multicolumn{5}{c}{Pole-like object positioning errors (m)} &  & \multicolumn{5}{c}{Building positioning errors (m)} \\ \cline{2-6} \cline{8-12} 
                         & N     & Min      & Max      & Avg.    & RMSE    &  & N      & Min       & Max      & Avg.     & RMSE     \\ \hline
1                        & 27    & 0.019    & 0.040    & 0.030   & 0.031   &  & 3      & 0.154     & 0.210    & 0.180    & 0.182    \\
2                        & 9     & 0.012    & 0.023    & 0.016   & 0.016   &  & 10     & 0.094     & 0.176    & 0.132    & 0.135    \\ \hline
\end{tabularx}
\end{table*}

\begin{figure*}[b!]
\centering
\begin{subfigure}[b]{0.49\linewidth}
    \centering
    \includegraphics[width=1\linewidth]{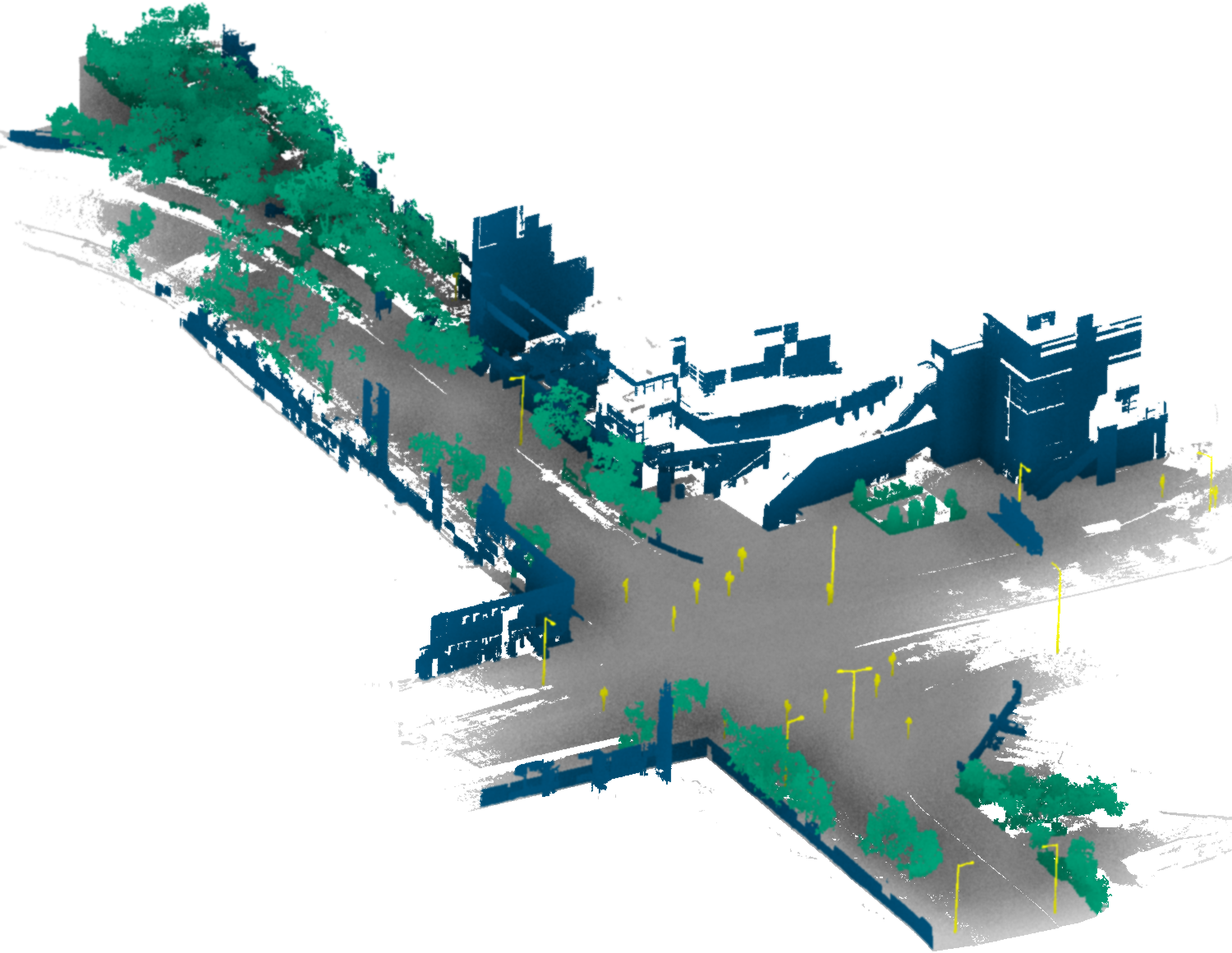}
    \caption{}
\end{subfigure}
\hfill
\begin{subfigure}[b]{0.49\linewidth}
    \centering
    \includegraphics[width=1\linewidth]{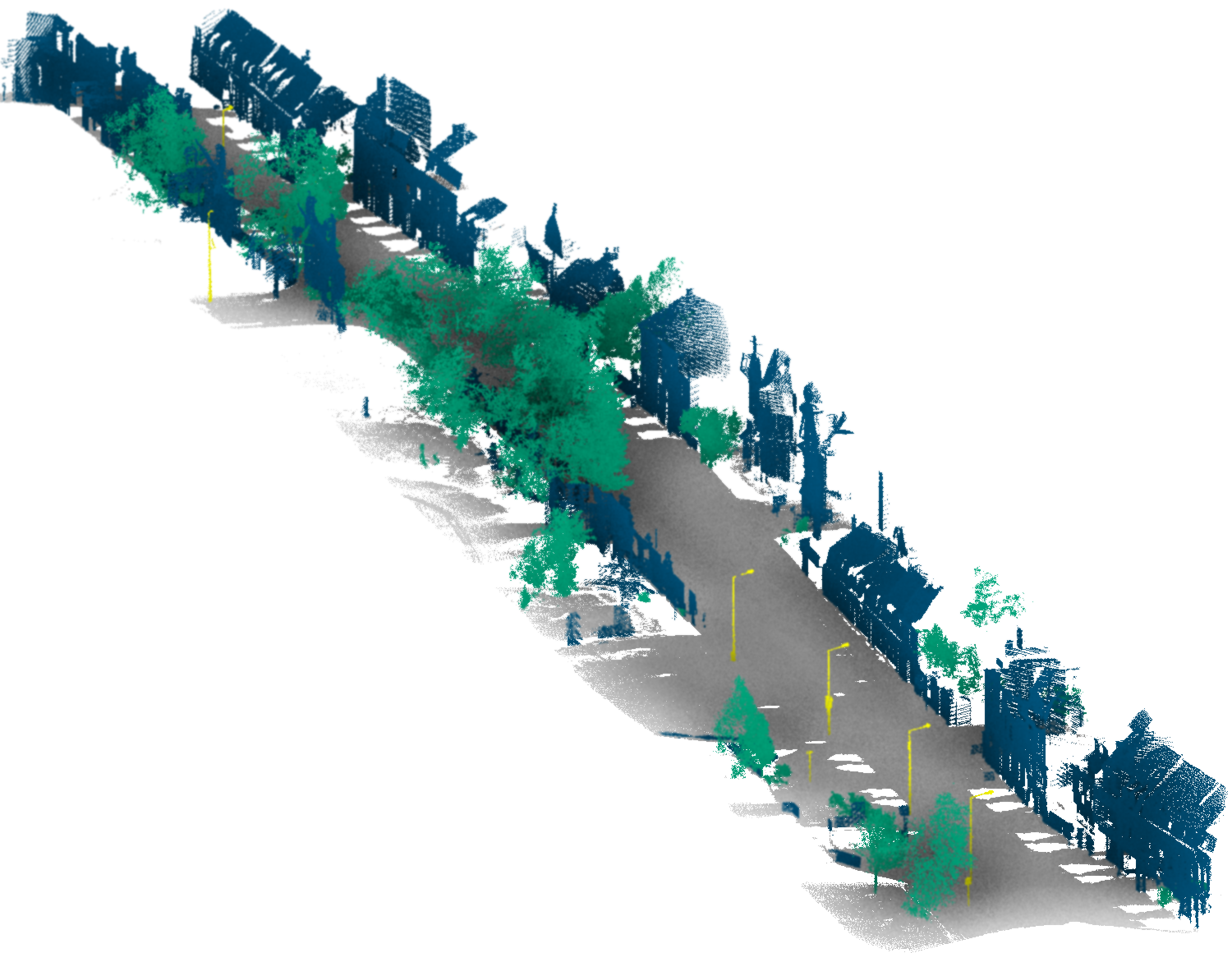}
    \caption{}
\end{subfigure}
\caption{Training samples labeling in Dataset 1 (a) and Dataset 2 (b).}
\label{fig:sample}
\end{figure*}

\subsection{Urban GeoBIM construction}
\label{sec:res_geobim}

Based on the instance segmentation process, individual pole-like objects and buildings can be obtained. Subsequently, the geo-referenced LiDAR point cloud is registered with the corresponding as-designed BIM model using the proposed matching strategy, with the LiDAR data serving as references. Once the LiDAR data and BIM models are matched, all as-designed BIM models will obtain geo-referenced information, resulting in the formation of urban GeoBIM models (Fig.~\ref{fig:geobim}).

The construction of urban GeoBIM models has resulted in a more accurate representation of 3D real cities in the form of geo-referenced BIM models. The spatial distribution of as-designed BIM models for all objects is consistent with that of their corresponding LiDAR point clouds, indicating the correctness and effectiveness of the matching process. To quantitively evaluate the accuracy of urban GeoBIM construction, the geo-referenced positions of all BIM models are compared with those of the corresponding LiDAR data. The position distances between BIM models and LiDAR data are calculated and presented in Fig.~\ref{fig:error} and Table~\ref{tab:error} for further analysis and assessment of the constructed GeoBIM models.

Positioning errors of building BIM models were generally larger than those of pole-like BIM models, which can be attributed to the large size and potential design errors in the building BIM model themselves. However, pole-like object positioning accuracies were high, with average errors and RMSE values less than 0.031 m in both datasets. This indicates the effectiveness of the coarse alignment and ICP-based fine registration algorithm for matching LiDAR point clouds and BIM models of pole-like objects. The building positioning accuracies varied between 0.1 m and 0.2 m, with buildings having complex structures generally exhibiting high positioning accuracy owing to rich structural information. Overall, the proposed method has the potential to construct accurate urban GeoBIM models.

\section{Discussions}
\label{sec:discuss}

\subsection{Analysis of training sample labeling}
\label{sec:dis_sample}

Training samples used in the supervised classification network are critical inputs that determine the effectiveness and accuracy of the semantic segmentation in this study. The results obtained so far have demonstrated the reliability of the training samples labeled by the proposed method. As shown in Fig.~\ref{fig:sample}, the training samples cover a wide range of urban scenes, and each category exhibited accurate labeling results. This not only provides reliable inputs for supervised semantic segmentation in an automatic manner but also takes into consideration the topological and geometric features at both global and local levels. This indicates that the proposed method takes into account various aspects of urban scenes, making it a robust and effective approach for semantic segmentation of LiDAR point clouds.

\begin{table}[b!]
\footnotesize
\caption{Labeling precision (\%) of training samples.}
\label{tab:sample}
\begin{tabularx}{\linewidth}{XXXXX}
\hline
\multirow{2}{*}{Dataset} & \multicolumn{4}{c}{Classes}            \\ \cline{2-5} 
                         & Ground & Building & Vegetation & Pole-like  \\ \hline
1                        & 94.58  & 82.03    & 85.81      & 98.65 \\
2                        & 98.51  & 83.08    & 96.47      & 96.79 \\ \hline
\end{tabularx}
\end{table}

\begin{figure*}[b!]
\centering
\begin{subfigure}[b]{0.49\linewidth}
    \centering
    \includegraphics[width=1\linewidth]{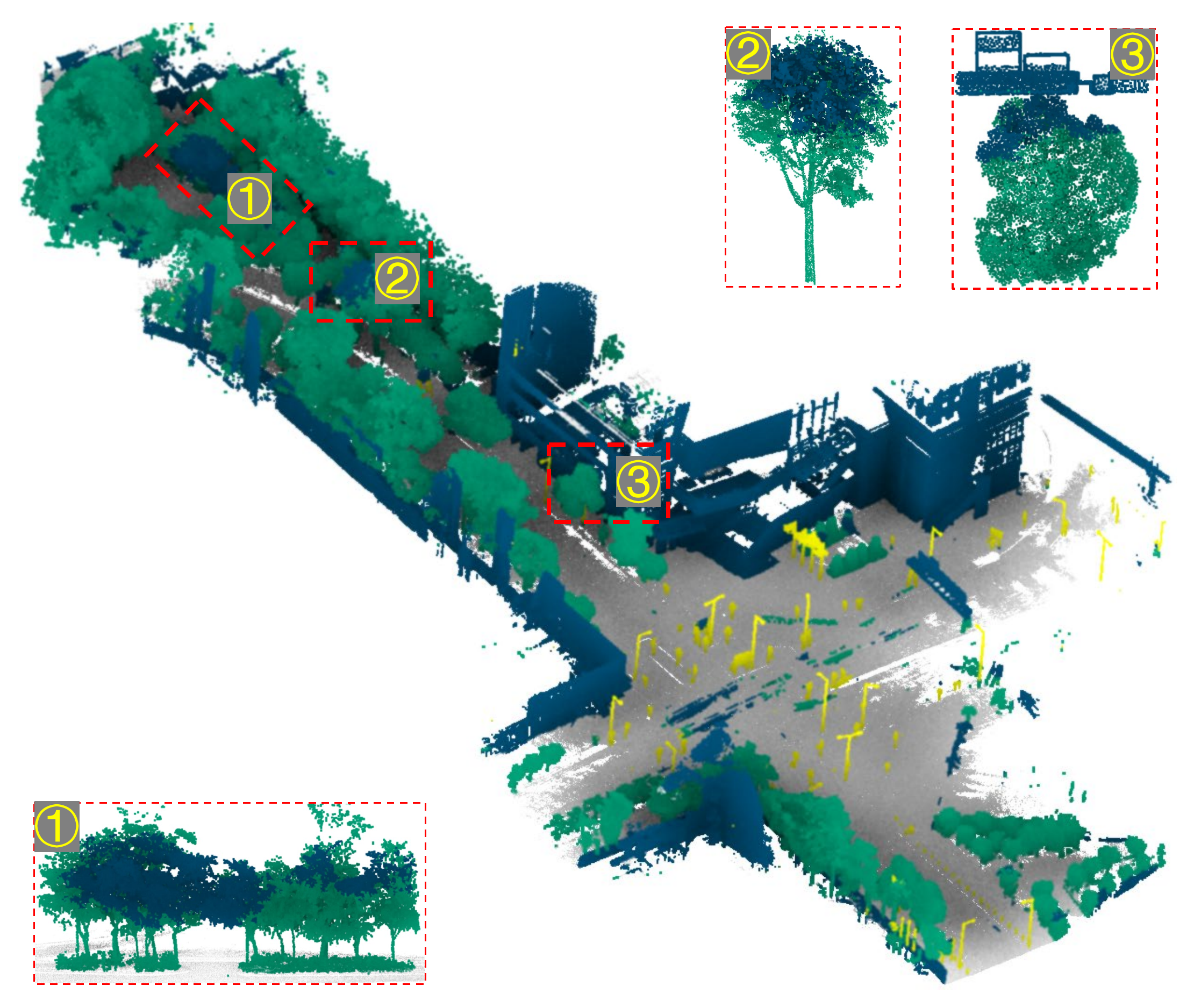}
    \caption{}
    \label{fig:ss_opt1}
\end{subfigure}
\hfill
\begin{subfigure}[b]{0.49\linewidth}
    \centering
    \includegraphics[width=1\linewidth]{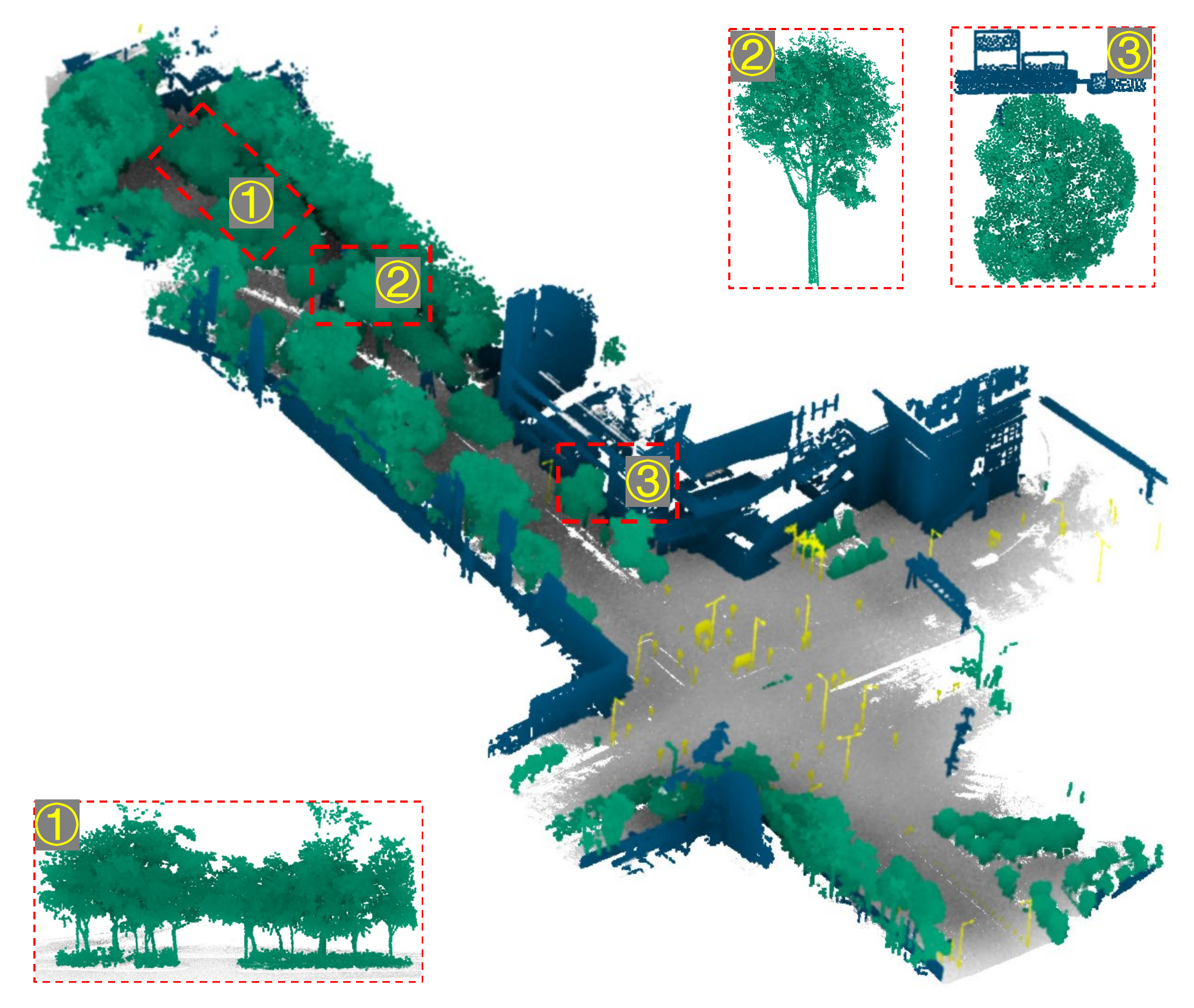}
    \caption{}
    \label{fig:ss_opt2}
\end{subfigure}
\caption{Segmentation optimization of Dataset 1, red rectangle in (a) shows some vegetation points (blue points) that were misclassified as buildings, and (b) shows the corrected results.}
\label{fig:cluster}
\end{figure*}

\begin{table*}[b!]
\footnotesize
\caption{The Confusion matrix of clustering results with per-class precision, recall, and F1 score (\%).}
\label{tab:cluster}
\begin{tabularx}{\textwidth}{XXXXXXXXX}
\hline
\multirow{2}{*}{Dataset} & \multirow{2}{*}{Criteria (\%)} & \multicolumn{5}{c}{Classes}                     & \multirow{2}{*}{Avg. F1 (\%)} & \multirow{2}{*}{OA (\%)} \\ \cline{3-7}
                         &                                & Others & Ground & Building & Vegetation & Pole-like  &                               &                          \\ \hline
\multirow{3}{*}{1}       & Precision                      & 99.08  & 96.75  & 77.62    & 97.81      & 98.52 & \multirow{3}{*}{86.96}        & \multirow{3}{*}{93.78}   \\
                         & Recall                         & 42.00  & 99.94  & 96.06    & 99.43      & 88.07 &                               &                          \\
                         & F1 score                       & 59.00  & 98.32  & 85.86    & 98.61      & 93.00 &                               &                          \\ \hline
\multirow{3}{*}{2}       & Precision                      & 87.61  & 99.08  & 82.06    & 89.13      & 98.71 & \multirow{3}{*}{77.81}        & \multirow{3}{*}{91.66}   \\
                         & Recall                         & 37.52  & 99.29  & 98.37    & 93.40      & 39.69 &                               &                          \\
                         & F1 score                       & 52.54  & 99.19  & 89.48    & 91.22      & 56.61 &                               &                          \\ \hline
\end{tabularx}
\end{table*}

The labeling precision of training samples is analyzed based on the criteria outlined in Section~\ref{sec:method_val} (Table~\ref{tab:sample}). It is found that nearly half of the LiDAR point cloud can be automatically labeled, providing support for the weakly supervised semantic segmentation network to obtain more reliable results. Furthermore, the labeling precision for each category is high, exceeding 80\% in the two datasets. Ground and pole-like categories exhibit particularly high labeling precision, reaching around 95\% owing to their unique geometric characteristics (elevation for ground and line features for pole-like objects). Buildings and vegetation categories can be generally distinguished based on their planar characteristics, with precisions of approximately 85\%. However, some irregular objects adjacent to vegetation, such as pedestrians and fences, may be mislabeled as the vegetation category due to varying the orientation of local surface points. Similarly, some regular objects adjacent to buildings, such as cars and tree trunks, may also be mislabeled as buildings due to local surface consistency characteristics. Nevertheless, the percentage of mislabeled points is small overall, and a large percentage of correctly labeled object points can ensure semantic segmentation accuracy.

\begin{figure*}[b!]
\centering
\begin{subfigure}[b]{0.55\linewidth}
    \centering
    \includegraphics[width=1\linewidth]{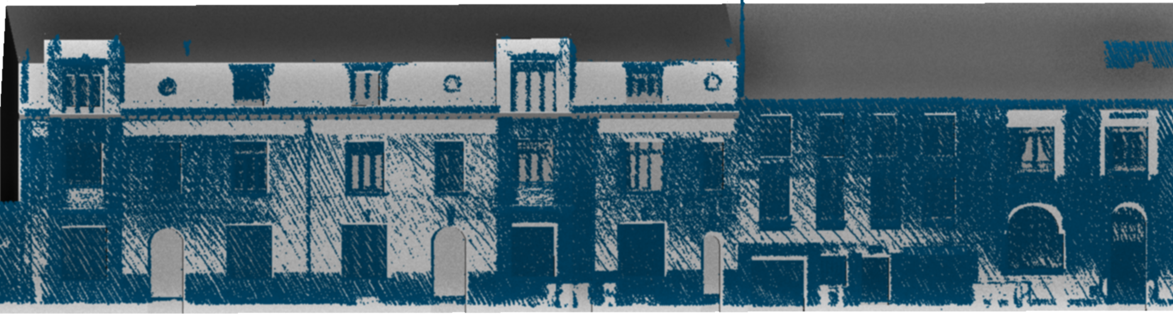}
    \caption{}
\end{subfigure}
\hfill
\begin{subfigure}[b]{0.43\linewidth}
    \centering
    \includegraphics[width=1\linewidth]{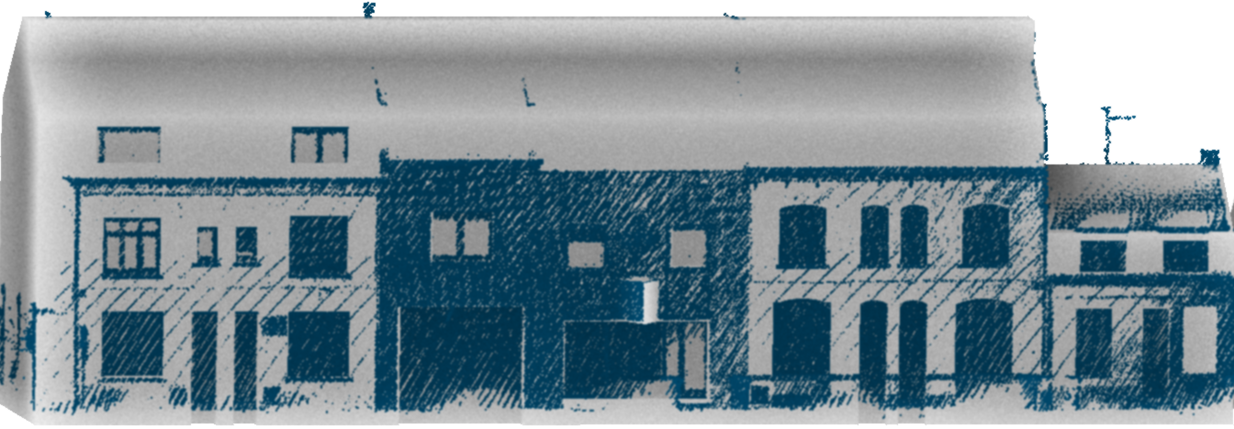}
    \caption{}
\end{subfigure}
\caption{Matching of BIM models (gray meshes) and LiDAR point clouds (blue points) of buildings in Dataset 2.}
\label{fig:match}
\end{figure*}

\begin{table*}[b!]
\footnotesize
\caption{Matching accuracy of buildings in Dataset 2. ‘N’ represents the number of correspondences.}
\label{tab:match}
\begin{tabularx}{\textwidth}{XXXXXXXXXXXXX}
\hline
\multirow{2}{*}{ID} & \multirow{2}{*}{N} & \multicolumn{4}{c}{errors (m)} &  & \multirow{2}{*}{ID} & \multirow{2}{*}{N} & \multicolumn{4}{c}{errors (m)} \\ \cline{3-6} \cline{10-13} 
                    &                    & Min    & Max   & Avg.  & RMSE  &  &                     &                    & Min    & Max   & Avg.  & RMSE  \\ \hline
1                   & 146                & 0.001  & 0.910 & 0.106 & 0.183 &  & 6                   & 65                 & 0.002  & 0.650 & 0.118 & 0.162 \\
2                   & 367                & 0.000  & 0.425 & 0.094 & 0.119 &  & 7                   & 405                & 0.000  & 0.777 & 0.135 & 0.188 \\
3                   & 45                 & 0.012  & 0.708 & 0.176 & 0.237 &  & 8                   & 161                & 0.002  & 0.811 & 0.120 & 0.175 \\
4                   & 30                 & 0.001  & 0.781 & 0.175 & 0.246 &  & 9                   & 507                & 0.000  & 0.580 & 0.124 & 0.180 \\
5                   & 112                & 0.000  & 0.544 & 0.112 & 0.161 &  & 10                  & 20                 & 0.005  & 0.412 & 0.164 & 0.200 \\ \hline
\end{tabularx}
\end{table*}

\subsection{Analysis of graph-based clustering}
\label{sec:dis_cluster}

Moreover, the proposed graph-based clustering method can also be used to optimize semantic segmentation results, in addition to its application in instance segmentation. In general, the weak supervision framework may produce false segmentation points due to the complexity of city scenes. For instance, vegetation points misclassified as buildings constitute the majority of false segmentation, and some internal points of buildings may be segmented as vegetation due to discrete characteristics (see Fig.~\ref{fig:cluster}).

The results of the proposed graph-based clustering method show that misclassified points, which are generally in the minority and adjacent to correctly segmented points, can be corrected using the distance information between adjacent objects, as seen in Fig.~\ref{fig:ss_opt1}. However, for points that are adjacent to multiple objects at the same time, it is challenging to accurately determine their final class based solely on distance information, as shown in Fig.~\ref{fig:ss_opt2}. To address this issue, the proposed clustering method integrates roughness and anisotropy information to optimize the incorrectly segmented points located at object adjacency, leveraging the similarity of local structural properties on the same object. 

Additionally, the method removes outliers that are characterized by discreteness and isolation, which effectively improves the visualization performance of semantic point clouds. As a result, the segmentation accuracy is improved compared to the semantic segmentation results, with a notable improvement of 3.3\% in overall accuracy and 1.96\% in average F1 score improved for Dataset 1, as shown in Table~\ref{tab:cluster}. This demonstrates the effectiveness of the graph-based clustering process in refining and improving the segmentation accuracy, particularly in cases where misclassified points are adjacent to multiple objects or characterized by discreteness and isolation.

\subsection{Matching performance evaluation}
\label{sec:dis_match}

The core of the proposed urban GeoBIM construction is to provide geo-referenced information for as-designed BIM models in the local coordinate system by aligning them with LiDAR point clouds. Therefore, the calculation of the transformation between BIM models and LiDAR point clouds becomes a key aspect of this work. While small-scale pole-like objects present relatively straightforward, large-scale buildings often pose more complex challenges due to their structural similarities. As a result, the experiments in this study focus on evaluating the matching results of as-designed BIM models and LiDAR point clouds of buildings, as shown in Fig.~\ref{fig:match}. The matching results demonstrate that BIM models can align well with their corresponding LiDAR point clouds, with the building facades in both datasets showing consistent alignment. This indicated that the matching results have the potential to provide accurate geo-referenced information for BIM models, ensuring precise positioning in urban scenes. To further quantify matching results, the distance between corners in BIM models and their corresponding planes in LiDAR data were calculated, as presented in Table~\ref{tab:match}.

The matching errors of all buildings were found to be similar, with average distance and RMSE values being less than 0.3 m. These errors may have been introduced due to various factors, such as the quality of BIM models and LiDAR data. In this work, BIM models were manually designed with reference to LiDAR data, which can be prone to human error and differences in LiDAR point cloud precision, resulting in inconsistencies between BIM models and LiDAR data, and subsequently affecting accurate correspondences. Additionally, LiDAR point clouds of incomplete buildings and sparse BIM model point clouds may result in low overlap, leading to inaccurate matching between LiDAR point clouds and BIM models. Therefore, if complete and high-precision LiDAR point clouds and as-designed BIM models were available, the matching accuracy could potentially be improved.

In general, ground-based LiDAR systems are capable of acquiring detailed structural information of urban scenes. However, due to limitations such as scanning location, the field of view, and measurement range, it is challenging for a single ground-based LiDAR system to capture complete urban scenes, resulting in incomplete building point clouds (Fig.~\ref{fig:data}). This incomplete data makes it difficult to directly construct detailed BIM models from LiDAR point clouds. To overcome this challenge, the most common approach currently used for data collection is to combine multiple platforms and utilize multi-source LiDAR point clouds, such as fusing airborne and ground-based LiDAR point clouds. However, compared to the acquisition of ground-based LiDAR data, acquiring airborne LiDAR point clouds is relatively knotty in urban scenes due to air restriction and safety factors. In modern infrastructure projects, detailed and precise BIM models are often designed prior to construction, and these rich BIM models can compensate for the lack of complete LiDAR data. It is in this context that we conducted this study.

\section{Conclusions}
\label{sec:conclusion}

In this paper, we propose a novel urban GeoBIM construction approach that leverages the complementary advantages of geo-referenced LiDAR point clouds and as-designed BIM models. The approach includes semantic segmentation, instance segmentation, and matching of LiDAR data and BIM models. In the semantic segmentation step, an automated training sample labeling method is presented for adaptively extracting ground, buildings, pole-like objects, and vegetation. Then, a graph-based clustering technique by combining topological and geometric features is used to optimize semantic segmentation results and identify instance buildings and pole-like objects. Furthermore, LiDAR point clouds and as-designed BIM models are matched through the coarse alignment and fine registration processes for generating urban GeoBIM models. Experimental results demonstrate that the proposed approach improves segmentation accuracies, particularly in Dataset 1 where the Avg. F1 score is improved by 1.96\%, from 85\% to 86.9\%. Moreover, the average positioning errors obtained by the matching method are 0.023 m for pole-like objects and 0.156 m for buildings, indicating the potential of achieving detailed and accurate urban GeoBIM construction.

The proposed method is verified in small streets with different urban characteristics to demonstrate its general applicability in urban 3D real scenes. Future work will focus on extending the approach to regional or urban scales for providing technical support and a comprehensive database for urban management and decision-making.

\section*{Declaration of Competing Interest}

The authors declare no conflict of interest.

\section*{Acknowledgements}

This work was supported by the National Natural Science Foundation of China (Project No.42171361) and the Research Grants Council of the Hong Kong Special Administrative Region, China, under Project PolyU 25211819. This work was also funded by the research project (Project Number: 2021.A6.184.21D) of the Public Policy Research Funding Scheme of The Government of the Hong Kong Special Administrative Region. This work was partially supported by The Hong Kong Polytechnic University under Projects 1-ZVN6, 1-YXAQ, and Q-CDAU.




\bibliographystyle{elsarticle-num} 
\bibliography{references}





\end{document}